\documentclass[11pt,a4paper]{article}
\usepackage[hyperref]{acl2020}
\usepackage{times}
\usepackage{latexsym}

\usepackage{mathtools}
\usepackage[utf8]{inputenc} % allow utf-8 input
\usepackage[T1]{fontenc}    % use 8-bit T1 fonts
\usepackage{hyperref}       % hyperlinks
\usepackage{url}            % simple URL typesetting
\usepackage{booktabs}       % professional-quality tables
\usepackage{graphicx}
\usepackage{float}
\usepackage{amsfonts}       % blackboard math symbols
\usepackage{nicefrac}       % compact symbols for 1/2, etc.
\usepackage{CJKutf8}
\AtBeginDvi{\input{zhwinfonts}}
\usepackage{amsmath}
\usepackage{adjustbox}
\usepackage[linesnumbered,ruled,vlined]{algorithm2e}
\usepackage{multicol}
\usepackage{arydshln}

\usepackage{multirow}
\usepackage[normalem]{ulem}
\useunder{\uline}{\ul}{}
\usepackage[american]{babel}
\usepackage{microtype}

\aclfinalcopy % Uncomment this line for the final submission
\title{Find or Classify? Dual Strategy for Slot-Value Predictions on Multi-Domain Dialog State Tracking}
\author{
  \textbf{Jian-Guo Zhang}$^1$\thanks{\ \ Work done while the first author was an intern at Salesforce Research.}~~~~ \textbf{Kazuma Hashimoto}$^2$\thanks{\ \ Corresponding author.}~~~~ \textbf{Chien-Sheng Wu}$^2$~~~~  \textbf{Yao Wan}$^3$  \\ \textbf{Philip S. Yu}$^1$~~~~  \textbf{Richard Socher}$^2$~~~~ \textbf{Caiming Xiong}$^2$ \\
  $^1$ University of Illinois at Chicago, Chicago, USA \\
  $^2$Salesforce Research,  Palo Alto, USA \\
  $^3$Huazhong University of Science and Technology, Wuhan, China\\
  \texttt{\{jzhan51,psyu\}@uic.edu}, \texttt{wanyao@hust.edu.cn} \\ \texttt{\{{k.hashimoto,wu.jason,rsocher,cxiong\}@salesforce.com}}
}
\date{}

\begin{document}
\maketitle
\begin{abstract}
Dialog state tracking (DST) is a core component in task-oriented dialog systems.
Existing approaches for DST mainly fall into one of two categories, namely, ontology-based and ontology-free methods.
An ontology-based method selects a value from a candidate-value list for each target slot, while an ontology-free method extracts spans from dialog contexts.
Recent work introduced a BERT-based model to strike a balance between the two methods by pre-defining categorical and non-categorical slots.
However, it is not clear enough which slots are better handled by either of the two slot types, and the way to use the pre-trained model has not been well investigated.
In this paper, we propose a simple yet effective dual-strategy model for DST, by adapting a single BERT-style reading comprehension model to jointly handle both the categorical and non-categorical slots.
Our experiments on the MultiWOZ datasets show that our method significantly outperforms the BERT-based counterpart, finding that the key is a deep interaction between the domain-slot and context information.
When evaluated on noisy (MultiWOZ 2.0) and cleaner (MultiWOZ 2.1) settings, our method performs competitively and robustly across the two different settings.
Our method sets the new state of the art in the noisy setting, while performing more robustly than the best model in the cleaner setting.
We also conduct a comprehensive error analysis on the dataset, including the effects of the dual strategy for each slot, to facilitate future research.

\end{abstract}

%%%%%%%%%%%%%%%%%%%%%%%%%%%%%%%%%

\section{Introduction}
\begin{figure}%[H]
    %\fbox{\rule{0pt}{3.0 in}
    \centering
    \resizebox{1.0\linewidth}{!}{%
    \centering
    \includegraphics[width=0.5\textwidth]{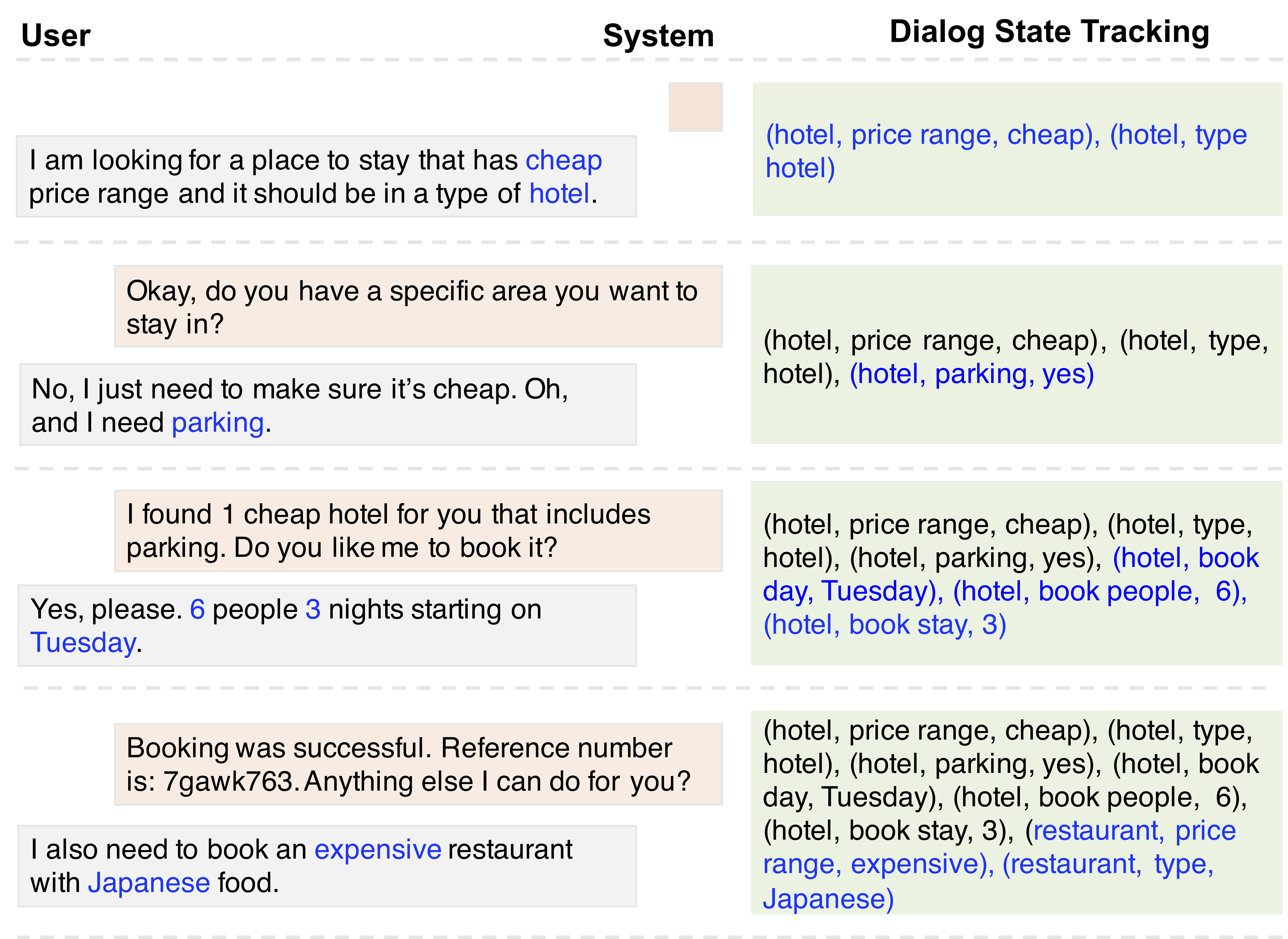}}
    %\rule{.9\linewidth}{0pt}}
    \caption{An example of dialog state tracking for booking a hotel and reserving a restaurant. Each turn contains a user utterance (grey) and a system utterance (orange). The dialog state tracker (green) tracks all the $<domain,slot, value>$ triplets until the current turn. Blue color denotes the new state appearing at that turn. Best viewed in color.}
    \label{fig:example}
    %\vspace{-1em}
\end{figure}

Virtual assistants play important roles in facilitating our daily life, such as booking hotels, reserving restaurants and making travel plans. 
% With the prevalence of virtual assistants such as Google Assistant, Cortana and Alexa, task-oriented dialog systems are playing important roles in facilitating our daily life, such as booking hotels, reserving restaurants and making traveling plans. 
% Dialog State Tracking (DST) is a core component in task-oriented dialog systems~\cite{young2013pomdp,gao2019neural}, which estimates users' goal and intention based on conversation history.
Dialog State Tracking (DST), which estimates users' goal and intention based on conversation history, is a core component in task-oriented dialog systems~\cite{young2013pomdp,gao2019neural}.
A dialog state consists of a set of $<domain, slot, value>$ triplets, and DST aims to track all the states accumulated across the conversational turns. Fig.~\ref{fig:example} shows a dialogue with corresponding annotated turn states.  

%To the best of our knowledge,
Traditional approaches for DST usually rely on hand-crafted features and domain-specific lexicon,  
and can be categorized into two classes~\cite{xu2018end, gao2019dialog,ramadan2018large,zhong2018global}:
%They usually fall into two categories
i.e., ontology-based and ontology-free. %picklist-based and span-based.
The ontology-based  approaches~\cite{ramadan2018large,zhong2018global,chen2020schema} require full access to the pre-defined ontology to perform classification over the candidate-value list.
% treat domain-slot pairs as picklist-based slots, where the values are predicted through performing classification on the candidate-value list. They usually require full access to the pre-defined ontology.
However, in practice, we may not have access to an ontology or only have partial ontology in the industry.
% would only have partial ontology since full ontology is hard and expensive to access in industry. 
Even if a full ontology exists, it is computationally expensive to enumerate all the values when the full ontology for some slots is very large and diverse~\cite{wu2019transferable,xu2018end}.
% , e.g., values for the \textit{time} slot could have unlimited choices.
The ontology-free approaches~\cite{gao2019dialog,xu2018end} find slot values directly from the input source using a copy mechanism without requiring an ontology, e.g., learning span matching with start and end positions in the dialog context.
% treat domain-slot pairs as span-based slots, where the values can be found through span matching with start and end positions in the dialog context. 
However, it is nontrivial to handle situations where values do not appear in the dialog context or have various descriptions by users.

To mitigate the above issues, recently, \cite{zhou2019multi} introduced a question asking model to generate questions asking for values of each-domain slot pair and a dynamic knowledge graph to learn relationships between the (domain, slot) pairs. \cite{rastogi2019towards} introduced a BERT-based model~\cite{devlin2018bert} to strike a balance between the two methods by pre-defining categorical and non-categorical slots. However, more studies are needed to know which slots are better handled by either of the two slot types, and the way to use the pre-trained models is not well investigated~\cite{lee2019sumbt,gao2019dialog,rastogi2019towards}.

% \textcolor{red}{Recently, some approaches~\cite{lee2019sumbt,gao2019dialog,rastogi2019towards} based on BERT~\cite{devlin2018bert} have been applied on multi-domain DST and have shown promising results on popular multi-domain dialog datasets, e.g., the MultiWOZ 2.1 dataset~\cite{eric2019multiwoz}. While they usually separately encode the dialog context and domain-slot pairs, it is interesting to study the influence of direct interactions between them. }

Inspired by the task-oriented dialog schema design in~\cite{rastogi2019towards} and the recent successful experience in locating text spans in machine reading comprehensions~\cite{gao2019dialog,asai}.
we design a simple yet effective \textbf{D}ual-\textbf{S}trategy \textbf{D}ialog \textbf{S}tate \textbf{T}racking model (\textbf{DS-DST}), which adapts a single BERT question answering model to jointly  handle  both  the  categorical  and  non-categorical slots, and different with previous approaches on multi-domain DST, we enable the model with direct interactions between dialog context and the slot.  We decide whether a slot belongs to a non-categorical slot or a categorical slot by following the heuristics from \cite{rastogi2019towards}.
For example, it is common that when users book hotels, the requests for parking are usually \textit{yes} or \textit{no} with limited choices.  These kinds of slots are defined as categorical slots, and the slot values are selected over a partial ontology. 
In addition, how long the user will stay has unlimited values and it can be found in the context. These kinds of slots are treated as non-categorical slots, and the values are found trough span matching in the dialog context. Hence, the model is flexible depending on the access level to the ontology or whether the values of slots could be found directly in the dialog context.

% \textcolor{red}{In this way, the model only needs partial ontology for picklist-based slots, and not for span-based slots. Moreover, it is flexible when the dialog system has access to the full ontology or all the values can be found in the dialog context, which also indicates that when the model has access to the partial ontology, these slots could be treated as picklist-based slots, otherwise these slots could be  handled by span-based slots. }

Our contributions are summarized as follows: 

$\bullet{}$ We designed a simple yet effective dual-strategy model based on BERT with strong interactions between the dialog context and domain-slot pairs. 

$\bullet{}$ Our model achieves state of the art on MultiWOZ 2.0~\cite{budzianowski2018multiwoz} and competitive performance on MultiWOZ 2.1~\cite{eric2019multiwoz}. Our model also performs robustly across the two different settings.

% $\bullet{}$ We investigated ways to use BERT and give suggestions to decide domain slot pairs.

$\bullet{}$ We conducted a comprehensive error analysis on the dataset, including  the  effects  of  the  dual  strategy  for each slot, to facilitate future research. 

% In this paper,
% % $\bullet{}$ 
% % we design an approach to treat domain-slot pairs as span-based slots and picklist-based slots.
% we first seperate the domain-slot pairs into picklist-based and span-based slots. Then, we propose a dual strategy to predict slot values via selecting over a partial ontology (for picklist-based) or finding values from the dialog context (for span-based slots). 
% Our approach mitigates the limitations of relying on fixed-vocabulary or unseen values of span and is flexible to different datasets and real scenarios.
% % $\bullet{}$
% We achieve state-of-the-art results on both MultiWOZ 2.0 dataset \cite{ramadan2018large} and MultiWOZ 2.1 dataset \cite{eric2019multiwoz}, and conduct comprehensive result analysis, \textcolor{red}{e.g. suggesting heuristics or ways to decide domain-slot types, and error analysis.}

\section{Related Work} \label{related-work}
% Dialog state tracking (DST) has been studied for several years, with the development of Dialogue State Tracking Challenges \cite{williams2013dialog,henderson2014second,henderson2014third,kim2017fourth,kim2016fifth,rastogi2019towards} and prevalence of conversational agents, it has becoming important that DST could track dialog state in complicated conversations and across multiple domains with many slots.
Multi-domain DST, which tracks dialog states in complicated conversations across multiple domains with many slots, has been a hot research topic during the past few years, along with the development of Dialogue State Tracking Challenges \cite{williams2013dialog,henderson2014second,henderson2014third,kim2016fifth,kim2017fourth,kim2019eighth}. 
Traditional approaches usually rely on hand-crafted features or domain-specific lexicon \cite{henderson2014word,wen2016network}, making them difficult to be adapted to new domains. In addition, these approaches require a pre-defined full ontology, in which the values of a slot are constrained by a set of candidate values \cite{ramadan2018large,liu2017end,zhong2018global,lee2019sumbt,chen2020schema}.
% However, these approaches are hard to handle unseen values and scale to large vocabularies. 
% Furthermore, these approaches are hard to adapt to unseen values and large vocabularies.
To tackle these issues, several methods have been proposed to extract slot values through span matching with start and end positions in the dialog context. For example,
\cite{xu2018end} utilizes an attention-based pointer network to copy values from the dialog context. 
\cite{gao2019dialog} poses DST as a reading comprehension problem and incorporates a slot carryover model to copy states from previous conversational turns.
However, tracking states only from the dialog context is insufficient since many values in DST cannot be exactly found in the context due to annotation errors or diverse descriptions of slot values from users. 
On the other hand, pre-trained models such as BERT~\cite{devlin2018bert} and GPT~\cite{radfordimproving} have shown promising performances in many downstream tasks. Among them, DSTreader~\cite{gao2019dialog} utilizes BERT as word embeddings for dialog contexts, SUMBT~\cite{lee2019sumbt} employs BERT to extract representations of candidate values, and BERT-DST~\cite{rastogi2019towards} adopts BERT to encode the inputs of the user turn as well as the previous system turn.
% Our model is different with them as our way to use BERT is more related to reading comprehensions \cite{chen2018neural}.  
Different from these approaches where the  dialog context and domain-slot pairs are usually separately encoded, we employ strong interactions to encode them. ~\footnote{Recent work on question answering has shown that the joint encoding of query-context pairs is crucial to achieving high accuracy~\cite{dfgn,asai}}. Moreover, We investigate and provide insights to decide slot types and conduct a comprehensive analysis of the popular MultiWOZ datasets. 
% \cite{wu2019transferable} designs a pointer generator to generate slot-values from utterances, which shows promising results on multi-domain DST. 

Another direction for multi-domain DST is based on generative approaches~\cite{lei2018sequicity,wu2019transferable,le2020non} which generate slot values without relying on fixed vocabularies and spans.
However, such generative methods suffer from generating ill-formatted strings (e.g., repeated words) upon long strings, which is common in DST. For example, the hotel address may be long  and a small difference makes the whole dialog state tracking incorrect.
By contrast, both the categorical (picklist-based) and non-categorical (span-based) methods can rely on existing strings rather than generating them.

\begin{figure*}
    %\fbox{\rule{0pt}{3.0 in}
    % \resizebox{1.0\textwidth}{!}{%
    \centering
    \includegraphics[width=0.98\textwidth]{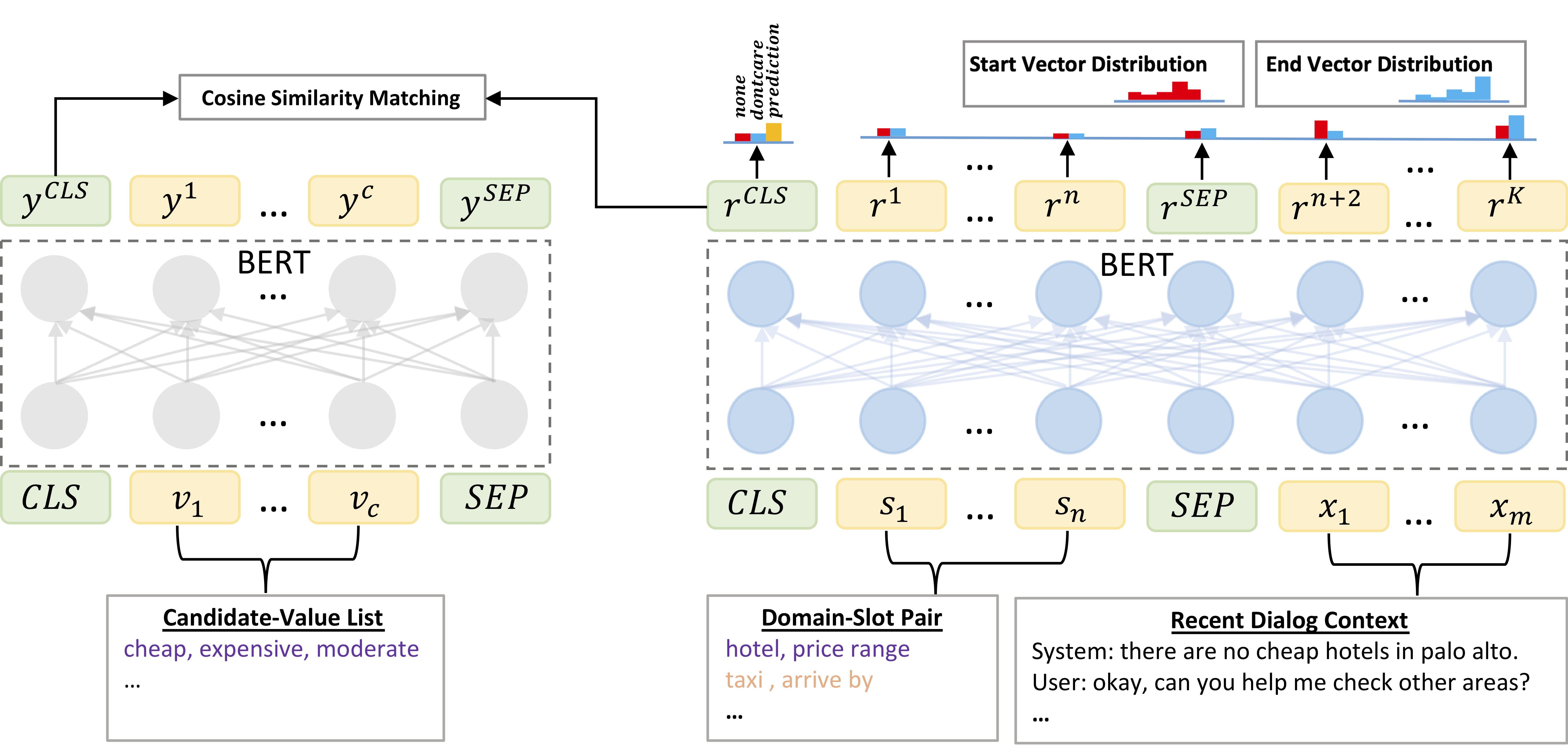}
    %\rule{.9\linewidth}{0pt}}
    \caption{The architecture of our proposed DS-DST model. The left part is a fixed BERT model which acts as a feature extractor and outputs the representations of  values in the candidate-value list for each categorical slot (marked in purple). The right part is the other fine-tuned BERT model which outputs representations for the concatenation of each domain-slot pair and the recent dialog context.
    % $\theta_{gate}$ is used for the slot-gate classification for all domain-slot pairs. $\theta_{span}$ is used to predict spans for each span-based domain-slot pair (orange). $\theta_{picklist}$ is used for cosine similarity matching over the candidate-value list for each picklist-based domain-slot pair (purple). Best viewed in color.
    }
    %}å
    \label{fig:framework1}
    \vspace{-0.5em}
\end{figure*}

\section{DS-DST: a Dual Strategy for DST}\label{bert-method}

Let $X=\left \{ (U_{1}^{sys},U_{1}^{usr}),\ldots,(U_{T}^{sys},U_{T}^{usr}) \right \}$ denote a set of pairs of a system utterance $U_{t}^{sys}$ and a user utterance $U_{t}^{usr}$ ($1 \leq t \leq T$), given a dialogue context with $T$ turns.
Each turn $(U_{t}^{sys}, U_{t}^{usr})$ talks about a particular domain (e.g., \textit{hotel}), and a certain number of slots (e.g., \textit{price range}) are associated with the domain.
We denote all the $N$ possible domain-slot pairs as $S=\left \{S_1,\ldots, S_N\right \}$, where each domain-slot pair consists of
$\left \{s_1,\ldots, s_n\right \}$ tokens, e.g., \textit{hotel-price range} includes three tokens.
% The goal of DST is to track the states over the whole dialogue. In other words, at each dialogue turn $t$, we need to predict the values for each $S$ (e.g., $<hotel, price \ range, cheap>$), considering the context $X_t = \left \{ (U_{1}^{sys},U_{1}^{usr}),\ldots,(U_{t}^{sys},U_{t}^{usr}) \right \}$, where $X_t$ has $m$ tokens.
% We follow recent strategies~\cite{wu2019transferable,xu2018end} to predict the values for all the domain-slot pairs in $S$ at each dialogue turn.
% The goal of DST is to track the states over the whole dialogue. In other words,
Let $X_t = \left \{ (U_{1}^{sys},U_{1}^{usr}),\ldots,(U_{t}^{sys},U_{t}^{usr}) \right \}$ denote the dialogue context at the $t_{th}$ turn and $X_t$ has $\left \{x_1,\ldots, x_m\right \}$ tokens. Our goal is to predict the values for all the domain-slot pairs in $S$.
%(e.g., $<hotel, price \ range, cheap>$). 
% Here we simply follow recent strategies which have been adopted in~\cite{wu2019transferable,xu2018end}.
% Our intuition is that we can find values from pre-defined picklists, if we have access to the partial or full candidate-value lists.
% Otherwise, we need to directly find the values as text spans in the dialogue context.
% We call the former type a picklist-based slot, and the latter one a span-based slot.
Here we assume that $M$ domain-slot pairs in $S$ are treated as non-categorical slots, and the remaining $N-M$ pairs as categorical slots.
Each categorical slot has $L$ possible candidate values (picklist), i.e., $\left \{V_1,\ldots, V_L\right \}$, where $L$ is the size of the picklist, and each value has $\left \{v_1,\ldots, v_c\right \}$ tokens.

Bearing these notations in mind, we then propose a dual strategy model with direct interactions between dialog context and domain-slot pairs for DST.
Fig.~\ref{fig:framework1} shows an overview of the architecture of our proposed DS-DST model.
We first utilize a pre-trained BERT~\cite{devlin2018bert} to encode information about the dialogue context $X_t$ along with each domain-slot pair in $S$, and obtain contextualized representations conditioned on the domain-slot information.
We then design a slot gate to handle special types of values.
In particular, for the non-categorical slots, we utilize a two-way linear mapping to find text spans. For the categorical slots, we select the most plausible values from the picklists based on the contextual representation.

% \textcolor{red}{In this paper, we follow the following heuristics or common sense, to decide which domain-slot pair is treated as either a span-based or picklist-based slot.}
% If a domain-slot pair can have the unlimited number of values, depending on different datasets, then it is treated as a span-based slot; otherwise, it is treated as a picklist-based slot.
% % i.e., it is not practical to enumerate all the \textit{time} and \textit{number} values,
%  For example, as the slot $arrive \ by$ of the $taxi$ domain is related to \textit{time}, its values could fall into $\left \{ 00:00-23:59\right \}$; as the slot $book \ stay$  of the $hotel$ domain is related to \textit{number}, its values belong to a constant range. Therefore, the \textit{time} and \textit{number} related slots are treated as span-based slots. On the other hand, for those slots which have a limited set of values, e.g., the slot $parking$ of the $hotel$ domain has three candidate values $\left \{yes, no, free\right \}$, we treat them as picklist-based slots. \textcolor{red}{Moreover, we will also investigate and suggest ways to decide slot types during the experimental discussions. }

\subsection{Slot-Context Encoder}
We employ a pre-trained BERT \cite{devlin2018bert} to encode the domain-slot types and dialog contexts.
For the $j_{th}$ domain-slot pair and the dialog context $X_t$ at the $t_{th}$ turn, we concatenate them and get corresponding representations:
\begin{equation}
    R_{tj}=\operatorname{BERT}\left ([\texttt{CLS}]\oplus S_j\oplus  [\texttt{SEP}] \oplus  X_t \right ),
\end{equation}\label{eq_equation1}
where \texttt{[CLS]} is a special token added in front of each sample, and \texttt{[SEP]} is a special separator token. The outputs of BERT in Eq.~(\ref{eq_equation1}) can be decomposed as $R_{tj}=[r_{tj}^{\texttt{CLS}},r_{tj}^{1},\ldots,r_{tj}^{K}]$, where $r_{tj}^{\texttt{CLS}}$ is the aggregated representation of the total $K$ sequential input tokens, and $[r_{tj}^{1},\ldots,r_{tj}^{K}]$ are the token-level representations. They are used for slot-value predictions in the following sections, and the BERT is fine-tuned during the training process.

\subsection{Slot-Gate Classification}\label{slot-gate-classify}

As there are many domain-slot pairs in multi-domain dialogues, it is nontrivial to correctly predict whether a domain-slot pair appears at each turn of the dialogue. Here we follow ~\cite{wu2019transferable,xu2018end} and design a slot gate classification module for our neural network. Specifically, at the $t_{th}$ turn, the classifier makes a decision among $\left \{none, dontcare, prediction\right \}$, where $none$ denotes that a domain-slot pair is not mentioned or the value is  `none' at this turn, $dontcare$ implies that the user can accept any values for this slot, and $prediction$ represents that the slot should be processed by the model with a real value. We utilize $r_{tj}^{\texttt{CLS}}$ for the slot-gate classification, and the probability for the $j_{th}$ domain-slot pair at the $t_{th}$ turn is calculated as:
\begin{equation}
     P_{tj}^{gate}=\operatorname{softmax}(W_{gate}\cdot \left ( r_{tj}^{\texttt{CLS}} \right )^\top +b_{gate}),
\end{equation}
% \begin{equation}
%     slot\_gate=argmax_i(p_{sg}^{i}), i\in \left \{ 1,2,3 \right \},
% \end{equation}
where $W_{gate}$ and $b_{gate}$ are learnable parameters and bias, respectively. 

% The loss for slot gate classification is computed as:
We adopt the cross-entropy loss function for the slot gate classification as follows:
\begin{equation}
    \mathcal{L}_{gate}=\sum_{t=1}^{T}\sum_{j=1}^{N}-\log(P_{tj}^{gate}\cdot (y_{tj}^{gate})^\top),
\end{equation}
where $y_{tj}^{gate}$ is the one-hot gate label for the $j_{th}$ domain-slot pair at the $t_{th}$ turn.

\subsection{Non-Categorical Slot-Value Prediction}
For each non-categorical slot, its value can be mapped to a span with start and end position in the dialog context, e.g., slot $leave \ at$ in the $taxi$ domain has spans $4:30$pm in the context. We take token-level representations $[r_{tj}^{1},\ldots,r_{tj}^{K}]$ of the dialog context as input, and apply a two-way linear mapping to get a start vector $\alpha_{tj}^{start}$ and an end vector $\alpha_{tj}^{end}$:
\begin{equation}
    \left [  \alpha _{tj}^{start},\alpha_{tj}^{end}\right ]=W_{span} \cdot \left ( [r_{tj}^{1},...,r_{tj}^{K}] \right )^\top+b_{span},
\end{equation}
where $W_{span}$ and $b_{span}$ are learnable parameters and bias, respectively. 
% After that we can get the probability distribution over all tokens for the $j_{th}$ slot:

The probability of the $i_{th}$ word being the start position of the span is computed as: 
$p_{tj}^{start_i}=\frac{e^{\alpha_{tj}^{start}\cdot r_{tj}^i}}{\sum_k \alpha_{tj}^{start}\cdot r_{tj}^k}$, and the loss for  the start position prediction can be calculated as:
\begin{equation}
    \mathcal{L}_{start}=\sum_{t=1}^{T}\sum_{j=1}^{M}-\log(P_{tj}^{start}\cdot (y_{tj}^{start})^\top),
\end{equation}
where $y_{tj}^{start}$ is the one-hot start position label for the $j_{th}$ domain-slot pair at the $t_{th}$ turn.

% the start position can be computed as:
% \begin{equation}
%     start\_pos=argmax_i(p_{start}^{i}), i\in \left \{ 1,...,K \right \}.
% \end{equation}
Similarly, we can also get the loss $\mathcal{L}_{end}$ for end positions prediction. Therefore, the total loss $\mathcal{L}_{span}$ for the non-categorical slot-value prediction is the summation of $\mathcal{L}_{start}$ and $\mathcal{L}_{end}$.

\subsection{Categorical Slot-Value Prediction}
Each categorical slot has several candidate values; e.g., slot $price \ range$ in the $hotel$ domain has three values $\left \{ cheap,expensive,moderate  \right \}$.
At the $t_{th}$ turn, for the $j_{th}$ domain-slot pair, we first use another pre-trained BERT to get the aggregated representation of each value in the candidate list:
\begin{equation}
    y_l^{\texttt{CLS}}=\operatorname{BERT}([\texttt{CLS}]\oplus V_l \oplus [\texttt{SEP}]),
\end{equation}
where $l\in \left \{1,\ldots,L  \right \}$.
% $Y_j=\left \{y_1^{\texttt{CLS}},...,y_L^{\texttt{CLS}}\right \}$, 
%$L$ is the number of candidate values.
Note that during the training process, this separate BERT model acts as a feature extractor and its model parameters are fixed. 

% We formulate a relevance score of the aggregated representation given a reference candidate by the cosine similarity \cite{lin2017adversarial}:
We calculate the relevance score between the aggregated representation and a reference candidate by the cosine similarity \cite{lin2017adversarial}:
\begin{equation}
   cos(r_{tj}^{\texttt{CLS}},y_l^{\texttt{CLS}})=\frac{r_{tj}^{\texttt{CLS}}\cdot (y_l^{\texttt{CLS}})^\top}{\left \| r_{tj}^{\texttt{CLS}} \right \| \left \| y_l^{\texttt{CLS}} \right \|},
\end{equation}
where $r_{tj}^{\texttt{CLS}}$ and $y_l^{\texttt{CLS}}$ are the aggregated representations from the slot-context encoder and the reference candidate value, respectively. 

During the training process, we employ a hinge loss to enlarge the difference between the similarity of $r_{tj}^{\texttt{CLS}}$ to the target value and that to the most similar value in the candidate-value list: 
% {\small
% \begin{equation}
% \begin{multlined}
%     \mathcal{L}_{picklist}=\sum_{t=1}^{T}\sum_{j=1}^{N-M} \max(0,\lambda-cos(r_{tj}^{\texttt{CLS}},y_{target}^{\texttt{CLS}}) \\ + \max_{y_{l}^{\texttt{CLS}}\neq y_{target}^{\texttt{CLS}}}cos(r_{tj}^{\texttt{CLS}},y_l^{\texttt{CLS}})),
% \end{multlined}
% \end{equation}
% }
\begin{align}
    \mathcal{L}_{picklist}&=\sum_{t=1}^{T}\sum_{j=1}^{N-M} \max(0,\lambda-cos(r_{tj}^{\texttt{CLS}},y_{target}^{\texttt{CLS}}) \nonumber \\
    	&+ \max_{y_{l}^{\texttt{CLS}}\neq y_{target}^{\texttt{CLS}}}cos(r_{tj}^{\texttt{CLS}},y_l^{\texttt{CLS}})), 
\end{align}
where $\lambda$ is a constant margin and $l\in \left \{1,\ldots,L  \right \}$, and $\mathcal{L}_{picklist}$ only requires partial ontology for DS-DST.

\subsection{Training Objective}
\label{training-objective}

During training process, the above three modules can be jointly trained and share parameters of BERT. We optimize the summations of different losses as:
\begin{equation}
    \mathcal{L}_{total}=\mathcal{L}_{gate}+\mathcal{L}_{span}+\mathcal{L}_{picklist}.
\end{equation}
For the slots that are not mentioned or the users can accept any values (i.e., slots $\in \left \{ none, dontcare \right \}$) at each dialogue turn, $\mathcal{L}_{span}$ and $\mathcal{L}_{picklist}$ are set to $0$ and only the slot-gate classification is optimized during the training process.

\section{Experimental Setup} \label{experiments-setting}
% In order to empirically evaluate the effectiveness of the proposed method in generating short product titles on E-commerce, we conduct experiments on a large scale data set and compare it with state-of-the-art methods. We further implement the framework in a real world online environment to test its practical performance. 

\begin{table}[]
\Huge
\resizebox{1.0\linewidth}{!}{%
\begin{tabular}{|l|c|c|c|c|c|c|c|}
\hline
\textbf{Domain}     & Hotel                                                                                                                                      & Train                                                                                                      & Restaurant                                                                                                    & Attraction                                                  & Taxi                                                                                    \\ \hline
\textbf{Slots}      & \begin{tabular}[c]{@{}c@{}}price range\\ type\\ parking\\ book stay\\ book day\\ book people\\ area\\ stars\\ internet\\ name\end{tabular} & \begin{tabular}[c]{@{}c@{}}destination\\ day\\ departure\\ arrive by\\ book people\\ leave at\end{tabular} & \begin{tabular}[c]{@{}c@{}}food\\ price range\\ area\\ name\\ book time\\ book day\\ book people\end{tabular} & \begin{tabular}[c]{@{}c@{}}area\\ name \\ type\end{tabular} & \begin{tabular}[c]{@{}c@{}}leave at \\ destination\\ departure\\ arrive by\end{tabular}                      \\ \hline
\textbf{Train}      & 3381                                                                                                                                       & 3103                                                                                                       & 3813                                                                                                          & 2717                                                        & 1654                                                                                \\ \hline %   & 8421                             & 56668                            \\ \hline
\textbf{Validation} & 416                                                                                                                                        & 484                                                                                                        & 438                                                                                                           & 401                                                         & 207                                                                           \\ \hline %           & 1001                             & 7374                             \\ \hline
\textbf{Test}       & 394                                                                                                                                        & 494                                                                                                        & 437                                                                                                           & 395                                                         & 195                                                                           \\ \hline %           & 1000                             & 7368                             \\ \hline
\end{tabular}
}
\caption{
% The dataset information of MultiWOZ 2.0 and MultiWOZ 2.1. There are in total 30 domain-slot pairs of five selected domains as shown in the top two rows. The last three rows show the number of dialogues for each  domain. 
% The last two columns show the total number of the dialogues and total conversational turns for all the domains. 
The dataset information of MultiWOZ 2.0 and MultiWOZ 2.1. 
The top two rows list 5 selected domains, consisting of 30 domain-slot pairs. The last three rows show the number of dialogues for each  domain.
}
 \label{table:domain-and-slot}
 %\vspace{-1em}
\end{table}

\subsection{Datasets}
% To demonstrate the performance of our DS-DST, 
% We use the multi-domain dataset MultiWOZ 2.0~\cite{budzianowski2018multiwoz} and its recent released version MultiWOZ 2.1~\cite{eric2019multiwoz}.\footnote{We also report the results on the single-domain WOZ 2.0 dataset \cite{wen2016network}, the dataset details and results can be found in the appendix.} MultiWOZ 2.0 is one of the largest multi-domain dialogue corpora with seven distinct domains and over $10,000$ dialogues. Compared with the original dataset, MultiWOZ 2.1 conducts dataset correction, including correcting dialog states, spelling errors, dialogue utterance corrections, and mis-annotations to reduce several substantial noises, making the dataset more challenging (more details can be found in \citet{eric2019multiwoz}). 

We use the  MultiWOZ 2.0~\cite{budzianowski2018multiwoz} dataset and the MultiWOZ 2.1~\cite{eric2019multiwoz} dataset. MultiWOZ 2.0 is one of the largest multi-domain dialogue corpora with seven distinct domains and over $10,000$ dialogues. Compared with the original dataset, MultiWOZ 2.1 conducts dataset correction, including correcting dialog states, spelling errors, dialogue utterance corrections, and mis-annotations to reduce several substantial noises (more details can be found in~\citet{eric2019multiwoz}). 

As \textit{hospital} and \textit{police} domains contain very few dialogues ($5\%$ of total dialogues), and they only appear in the training dataset, we ignore them in our experiments, following~\citet{wu2019transferable}. We adopt only five domains (i.e., \textit{train}, \textit{restaurant}, \textit{hotel}, \textit{taxi}, \textit{attraction}) and obtain totally $30$ domain-slot pairs in the experiments. Table~\ref{table:domain-and-slot} summarizes the domain-slot pairs and their corresponding statistics in MultiWOZ 2.0 and MultiWOZ 2.1.
% We utilize the standard training/validation/test shown in the dataset.
We follow the standard training/validation/test split strategy provided in the original datasets, and the data pre-processing script provided in~\citet{wu2019transferable}. 
%Instead of formulating the candidate-value list for each picklist-based slot through directly using the incomplete ontology~\cite{lee2019sumbt} of MultiWOZ 2.0 and MultiWOZ 2.1, we construct the candidate-value list for each picklist-based slot through traversing the dataset.

For MultiWOZ 2.0 and 2.1, the candidate values for the categorical slots are derived based on the ground-truth values of each slot that appeared in the partial dataset ontology. Besides, Since there are no provided ground-truth start positions and end positions for non-categorical slots in the datasets, we find the spans trough string matching between the ground truth values and the values in the dialog contexts, and we treat the start and end positions of the span which appeared at the most recent dialog turn as the ground-truth start positions and end positions.

\begin{table*}[]
\centering
%\resizebox{0.9\linewidth}{!}{
{
\begin{tabular}{llll}
\hline
\textbf{Models}       & \textbf{\begin{tabular}[c]{@{}l@{}}MultiWOZ 2.0\\ \end{tabular}} & \textbf{\begin{tabular}[c]{@{}l@{}}MultiWOZ 2.1\\ \end{tabular}} \\ \hline
SpanPtr   \cite{xu2018end}         & 30.28\%                                                                     & 29.09\%                                                                     \\
Ptr-DST        &- & 42.17\%                 \\
DSTreader \cite{gao2019dialog}  & 39.41\%                                                                     & 36.40\%$^{\star}$                                                                     \\
% DSTreader (single) \cite{gao2019dialog}  & 39.41\%                                                                     & 36.40\%$^{\star}$                                                                     \\
%DSTreader (ensemble)~\cite{gao2019dialog} & 42.12\%                                                                     & -                                                                           \\
TRADE   \cite{wu2019transferable}          & 48.62\%                                                                     & 45.60\%$^{\star}$                                                                     \\
%SUMBT$^{+}$ \cite{lee2019sumbt} & 46.65\% & - \\ 
COMER   \cite{ren2019scalable}         & 45.72\%                                                                     & -                                                                           \\
DSTQA w/span   \cite{zhou2019multi}         & 51.36\%                                                                     & 49.67\%
\\
DSTQA w/o span$\mathbf{^+}$   \cite{zhou2019multi}         & 51.44\%                                                                     & 51.17\%                                                                           \\
BERT-DST \cite{rastogi2019towards}        & -                                                                    & 43.40\%                                                                     \\ 
MA-DST \cite{kumar2020ma}        & -                                                                    & 51.04\%                                                                     \\ 

SST-2$\mathbf{^+}$  \cite{chen2020schema}       & 51.17\%                                                                   & \textbf{55.23\%}                                                                     \\ 
%MD-DST (ensemble) \cite{rastogi2019towards}        & -                                                                   & 51.88\% \\
NA-DST \cite{le2020non} & 50.52\% & 49.04\% 
\\ \hline
DS-Span             & \multicolumn{1}{l}{42.59\%}                                                        & \multicolumn{1}{l}{40.00\%}                                                        \\
DS-DST               & \multicolumn{1}{l}{52.24\%}                                                        & \multicolumn{1}{l}{51.21\%}                                                       \\ 
DS-Picklist$\mathbf{^+}$         & \multicolumn{1}{l}{\textbf{54.39\%}}                                                        &  \multicolumn{1}{l}{53.30\%}                                                        \\ \hline
\end{tabular}}
\caption{Joint accuracy on the test sets of MultiWOZ 2.0 and  2.1. $\mathbf{^+}$: the models require a full ontology, and $^{\star}$: the results are reported by \citet{eric2019multiwoz}} \label{table-jointacc}
\vspace{-0.5em}
\end{table*}
%%%%%%%%%%%%%%%%%%%%%%%%%%%%%%%%%%%%%%%%%%%%%%%%%%%%%%%%%%%%%%%%%%

\subsection{Models}
% Due to the lack of official pre-processing standard for the MultiWOZ dataset,\footnote{The pre-processing way provided in \cite{wu2019transferable} is currently suggested by the data providers:  \url{http://dialogue.mi.eng.cam.ac.uk/index.php/corpus/}} different ways would affect the performance evaluation (more details can be found in Sec. \ref{discussion}). 
% To make a  fair comparison, we only adopt several state-of-the-art baselines which either follow the same data pre-processing way \cite{wu2019transferable} or the results are provided by the data providers \cite{eric2019multiwoz}: 

We make a comparison with several existing models \footnote{We did not compare with \cite{lee2019sumbt} and \cite{shan2020contextual} as the datasets preprocessing is different with other baselines and ours.} and introduce some of them as below:

$\bullet{}$ \textbf{SpanPtr} \cite{xu2018end}. It applies a RNN-based pointer network  to find text spans with start and end pointers for each domain-slot pair.

$\bullet{}$ \textbf{Ptr-DST}. It is a variant based on SpanPtr with the exception that some slots are categorical slots, following DS-DST. %and can be found in candidate-value
% $\bullet{}$ \textbf{GLAD} \cite{zhong2018global}: It uses global self-attentive modules to share parameters among different slots and local modules to learn slot-specific feature representations.

% $\bullet{}$ \textbf{FJST} and \textbf{HJST} \cite{eric2019multiwoz}: FJST contains a bidirectional LSTM network to encode the dialog context and a separate feedforward network to predict each dialog state slot. HJST uses a similar architecture to FJST but encodes the history using a hierarchical recurrent network.

% $\bullet{}$ \textbf{HyST} \cite{goel2019hyst}. It is a hybrid approach  based on hierarchical RNNs and an open-vocabulary generator.

$\bullet{}$ \textbf{DSTreader} \cite{gao2019dialog}. It models the DST from the perspective of machine reading comprehensions and applies a pre-trained BERT as initial word embeddings. 

$\bullet{}$ \textbf{DSTQA} \cite{zhou2019multi}. It applies a dynamically-evolving knowledge graph and generates question asking for  the values of a domain-slot pair.

$\bullet{}$ \textbf{TRADE} \cite{wu2019transferable}. It contains a slot gate module for slots classification and a pointer generator for states generation. 

% $\bullet{}$ \textbf{SUMBT} \cite{lee2019sumbt}. It applies BERT as the encoder of utterances and minimizes the distance between the dialog context output vector and the slot-value's semantic vectors. 

$\bullet{}$ \textbf{COMER} \cite{ren2019scalable}. It applies BERT as contextualized word embeddings and first generates the slot sequences in the belief state, then generates the value sequences for each slot.
% $\bullet{}$ \textbf{COMMER} \cite{ren2019scalable}: It is a conditional memory relation network which formulates the dialogue state tracking as a sequential generation problem. It applies BERT as contextualized word embeddings and first generates the slot sequences in the belief state, then generate the value sequences for each slot.

$\bullet{}$ \textbf{BERT-DST} \cite{rastogi2019towards}. It uses BERT to obtain schema element embeddings and encode system as well as user utterances for dialogue state tracking. Different from the original model, it incorporates a pointer-generator copying mechanism for non-categorical slots of the MultiWOZ datasets. 

$\bullet{}$ \textbf{SST-2} \cite{chen2020schema}. It uses the graph neural network to incorporate slot relations and model slot interactions.

% $\bullet{}$ \textbf{MA-DST}\cite{kumar2020ma}.  It applies
% cross-attention and self-attention to model cross-domain relationships between the context and slots under different semantic levels.

% $\bullet{}$ \textbf{NA-DST} \cite{le2020non}. It is a  non-autoregressive model to learn inter-dependencies across slots for decoding dialogue states and capture dependencies at token level with low latency.

For our proposed methods, we design three variants: 

$\bullet{}$ \textbf{DS-DST}. It represents our proposed dual strategy model for DST, which can simultaneously handle the non-categorical slots as well as the categorical ones. Following heuristics from \cite{rastogi2019towards}, \textit{time} and \textit{number} related slots are treated as non-categorical slots, resulting in five slot types across four domains (nine domain-slot pairs in total), and the rest slots are treated as categorical slots (See also in Table \ref{table:per-slot-acc}). We also conduct investigations to decide domain-slot types in the experiments. 
% For the MultiWOZ 2.0 and MultiWOZ 2.1, \textit{time} and \textit{number} related slots are treated as span-based slots, resulting in  five slot types across four domains (nine domain-slot pairs in total), and the rest slots are treated as picklist-list based slots. 

$\bullet{}$ \textbf{DS-Span}. Similar to \citet{xu2018end,gao2019dialog}, it treats all domain-slot pairs as non-categorical slots, where corresponding values for each slot are extracted through text spans (string matching) with start and end positions in the dialog context. 

$\bullet{}$ \textbf{DS-Picklist}. Similar to \cite{lee2019sumbt,chen2020schema}, It assumes a full ontology is available and treats all domain-slot pairs as categorical slots, where corresponding values for each slot are found in the candidate-value list (picklist). 

% $\bullet{}$ \textbf{DS-DST}. It represents our proposed dual strategy for DST, which can simultaneously handle the span-based slots as well as the picklist-based ones. For the MultiWOZ 2.0 and MultiWOZ 2.1, \textit{time} and \textit{number} related slots are treated as span-based slots, resulting in  five slot types across four domains (nine domain-slot pairs in total), and the rest slots are treated as picklist-list based slots. 

% $\bullet{}$ \textbf{Span-DST}. Similar to \citet{xu2018end}, it treats all domain-slot pairs as span-based slots, where corresponding values for each slot are extracted through text spans (string matching) with start and end positions in the dialog context. 

% $\bullet{}$ \textbf{Picklist-DST}. It assumes a full ontology is available and treats all domain-slot pairs as picklist-based slots, where corresponding values for each slot are found in the candidate-value list. 

\section{Experimental Results} \label{expriment-results}
We evaluate all the models using the joint accuracy metric.
At each turn, the joint accuracy is $1.0$ if and only if all $<domain, slot, value>$ triplets are predicted correctly, otherwise $0$.
The score is averaged across all the turns in the evaluation set.

\subsection{Joint Accuracy}
% \textcolor{blue}{
\paragraph{Overall performance}

Table~\ref{table-jointacc} shows the results on the test sets of two datasets. 
We can see that our models achieve the top performance on MultiWOZ 2.0 and competitive performance on MultiWOZ 2.1.
% which surpasses the current state-of-the-art, TRADE, by $3.62\%$ on MultiWOZ 2.0. Moreover,
% the model improves the performance by  $5.61\%$ on the more challenging MultiWOZ 2.1. 
Among these state-of-the-art results, ours are less sensitive to the dataset differences. 

% \begin{table}[]
% \centering
% \resizebox{0.9\linewidth}{!}{
% \begin{tabular}{lll}
% \hline
% \textbf{Models} & \textbf{MultiWOZ 2.0} & \textbf{MultiWOZ 2.1} \\ \hline
% Span-DST        & 42.59\%               & 40.00\%               \\
% DS-DST          & 52.24\%               & 51.21\%               \\
% Picklist-DST    & 54.39\%               & 53.30\%               \\ \hline
% \end{tabular}}\caption{\textcolor{blue}{Joint accuracy of variations on the test sets of MultiWOZ 2.0 and 2.1.}}\label{table-span-picklist}
% \end{table}

\begin{table}[]
\centering
\resizebox{1.0\linewidth}{!}{
\begin{tabular}{ll}
\hline
\textbf{Models}                          & \textbf{Joint Accuracy} \\ \hline
BERT-DST   \cite{rastogi2019towards}                              & 43.40\%                 \\ 
DS-DST                                   & 51.21\%                 \\ \hdashline
BERT-DST-Picklist (single turn)          & 39.86\%                 \\
BERT-DST-Picklist (whole dialog history) & 46.42\%                 \\
ToD-BERT \cite{wu2020tod} & 48.00\%                 \\
DS-Picklist                                   & 53.30\%                 \\ \hline
\end{tabular}}\caption{Joint accuracy on the test sets of MultiWOZ 2.1. BERT-DST is the model used in MultiWOZ 2.1. BERT-DST-Picklist is the original model described in \cite{rastogi2019towards}, where a full ontology is required and  all the slots are treated as categorical slots,. `single turn' and `whole dialog history' represent the Bert utterance inputs are the current dialog turn and the whole dialog history, respectively.} \label{table-compare-with-google}
%\vspace{-1em}
\end{table}

%%%%%%%%%%%%%%%%%%%%%%%%%%%%%%%%%%%%%%%%%%%
\begin{table}[t]
\centering
\Huge
%\resizebox{1.0\linewidth}{!}{
\begin{adjustbox}{width=1\linewidth}
\begin{tabular}{llll}
\hline
\textbf{Slot Name}     & \textbf{DS-Span} & \textbf{DS-DST} & \textbf{DS-Picklist} \\ \hline
hotel-type             & 87.92            & 93.97 (\textbf{+6.05})   & 94.29 (\textbf{+6.37})        \\
attraction-name        & 91.16            & 93.81 (\textbf{+2.65})   & 93.93 (\textbf{+2.77})        \\
restaurant-name        & 92.11            & 93.38 (+1.27)   & 92.89 (+0.78)        \\
hotel-internet         & 92.98            & 97.48 (\textbf{+4.50})   & 97.26 (\textbf{+4.28})        \\
hotel-parking          & 93.42            & 97.18 (\textbf{+3.76})   & 96.99 (\textbf{+3.57})        \\
attraction-type        & 93.77            & 96.86 (\textbf{+3.09})   & 96.91 (\textbf{+3.14})        \\
hotel-name             & 94.19            & 94.87 (+0.68)   & 94.77 (+0.58)        \\
hotel-area             & 94.73            & 95.87 (+1.14)   & 95.47 (+0.74)        \\
restaurant-area        & 96.23            & 96.86 (+0.63)   & 97.18 (\textbf{+0.95})        \\
attraction-area        & 96.57            & 96.96 (+0.39)   & 96.73 (+0.16)        \\
hotel-price range      & 96.92            & 97.39 (+0.47)   & 96.97 (+0.05)        \\
train-departure        & 96.96            & 98.55 (\textbf{+1.59})   & 98.34 (\textbf{+1.38})        \\
restaurant-food        & 97.24            & 97.60 (+0.36)   & 97.19 (-0.05)        \\
restaurant-price range & 97.29            & 97.73 (+0.44)   & 97.69 (+0.40)        \\
taxi-departure         & 97.57            & 98.53 (\textbf{+0.96})   & 98.59 (\textbf{+1.02})        \\
taxi-destination       & 97.69            & 98.49 (\textbf{+0.80})   & 98.24 (+0.55)        \\
hotel-stars            & 97.80            & 97.48 (-0.32)   & 97.76 (-0.04)        \\
train-destination      & 98.17            & 98.86 (\textbf{+0.69})   & 98.59 (+0.42)        \\
train-day              & 99.24            & 99.35 (+0.11)   & 99.33 (+0.09)        \\
hotel-book day         & 99.40            & 99.32 (-0.08)   & 99.24 (-0.16)        \\
restaurant-book day    & 99.40            & 99.57 (+0.17)   & 99.44 (+0.04)        \\ \hdashline
train-leave at         & 93.43            & 93.30 (-0.13)   & 93.91 (+0.48)        \\
train-arrive by        & 95.25            & 95.78 (+0.53)   & 96.59 (\textbf{+1.34})        \\
train-book people      & 97.99            & 97.84 (-0.15)   & 98.51 (+0.52)        \\
restaurant-book time   & 98.56            & 98.44 (-0.12)   & 99.04 (+0.48)        \\
taxi-leave at          & 98.63            & 98.53 (-0.10)   & 98.94 (+0.31)        \\
hotel-book people      & 99.06            & 99.04 (-0.02)   & 99.29 (+0.23)        \\
taxi-arrive by         & 99.12            & 99.01 (-0.11)   & 99.09 (-0.03)        \\
hotel-book stay        & 99.25            & 99.25 (+0.00)   & 99.40 (+0.15)        \\
restaurant-book people & 99.31            & 99.16 (-0.15)   & 99.44 (+0.13)        \\ \hdashline
Average Accuracy & 96.38\% & 97.35\% & 97.40\% \\ \hline
\end{tabular}
%\vspace{-1em}
%}
\end{adjustbox}
\caption{The slot-level accuracy on the test set of MultiWOZ 2.1.
`+/-' indicates absolute performance improvement/degradation compared with DS-Span.
The numbers highlighted in bold indicate that the difference is significant ($p<0.05$), tested by bootstrap re-sampling~\citep{bootstrap}.
The slots above the first dashed line are categorical slots and the slots below the first dashed line are non-categorical slots for DS-DST. The last row shows the average slot accuracy.} \label{table:per-slot-acc}
%\vspace{-1em}
\end{table}

% Table \ref{table-span-picklist} shows the results of variations.   Comparing DS-Span and DS-DST, we can find that jointly using the span-based and picklist-based approaches is indeed helpful in multi-domain DST. When the model has access to the full ontology, Picklist-DST shows that our method could further improve the DST performance. Although Picklist-DST is higher than DS-DST, in real scenarios, it may be nontrivial to have access to the full ontology. 

Comparing DS-Span and DS-DST, we can find that jointly using the non-categorical and categorical approaches is indeed helpful in multi-domain DST. When the model has access to the full ontology, DS-Picklist shows that our method could further improve the DST performance. Although DS-Picklist is higher than DS-DST, in real scenarios, it may be nontrivial to have access to the full ontology. In the paper, we jointly train the three modules in Section \ref{training-objective}, we also conduct experiments for separately training the non-categorical slots and categorical slots. DS-DST drops by $1.90\%$ on MultiWOZ 2.1, which shows the benefits of jointly training.

%%%%%%%%%%%%%%%%%%%%%%%%%%%%%%%%%%%%%%%%%%%%%%
\begin{table*}[]
\centering
%\resizebox{0.82\linewidth}{!}{
{
%\begin{adjustbox}{width=\textwidth}
\begin{tabular}{lcrr}
\hline
\textbf{Slot Name} & \textbf{\begin{tabular}[c]{@{}c@{}}DS-Span\\ (\textit{\#Unfound / \#Relative\_Turns})\end{tabular}} & \textbf{DS-DST} & \textbf{DS-Picklist} \\ \hline
hotel-type         & 667/1395                                         & 86.36\%                                                                           & 85.91\%                                                                                \\
hotel-parking      & 419/1048                                         & 89.50\%                                                                           & 86.63\%                                                                                \\
hotel-internet     & 421/1124                                         & 95.72\%                                                                           & 94.54\%                                                                                \\
taxi-leave at      & 73/364                                           & 0.00\%                                                                          & 43.84\%                                                                                  \\
attraction-name    & 215/1261                                         & 70.23\%                                                                           & 74.42\%                                                                                \\
attraction-type    & 270/1658                                         & 84.81\%                                                                           & 84.07\%                                                                                \\
train-leave at     & 181/1164                                         & 2.21\%                                                                             & 41.44\%                                                                                 \\
hotel-area         & 168/1452                                         & 51.19\%                                                                           & 58.93\%                                                                                 \\
train-arrive by    & 125/1428                                         & 9.60\%                                                                            & 79.20\%                                                                                 \\
attraction-area    & 177/1620                                         & 67.23\%                                                                           & 71.75\%                                                                                \\ \hline
\end{tabular}}
%\end{adjustbox}
\caption{Statistics of Top-10 slots on the MultiWOZ 2.1 validation set based on (\textit{\#Unfound / \#Relative\_Turns}). DS-DST and DS-Picklist show percentages based on (\textit{\#Recover / \#Unfound}). \textit{\#Unfound} is the number of slots whose values cannot be found through span matching in the dialog context,  \textit{\#Relative\_Turns} is the number of dialogue turns where the slot type is mentioned, and \textit{\#Recover} indicates the number of values correctly predicted by DS-DST or DS-Picklist.}
\label{table-error-analysis}
%\vspace{-1em}
\end{table*}

\paragraph{Detailed comparisons with BERT related methods}

Compared with those methods as shown in Table \ref{table-jointacc}, we can observe that DS-Span, which employs the strength of BERT, outperforms SpanPtr by $10.91\%$, and it outperforms COMMER and DSTreader, which also use a pre-trained BERT model as dialog context embeddings and word embeddings, respectively. 
% These results show that it is effective to use BERT as a well-established machine reading comprehension model~\citep{devlin2018bert} for dialogue understanding. 
% DS-DST outperforms DS-Span by $11.21\%$, and Ptr-DST outperforms SpanPtr by $13.08\%$ on the MulitiWOZ 2.1, which implies that our model using both the span-based and picklist-based approaches could significantly improve the DST performance when tracking states across different domains.
% We attribute the performance difference between the dual strategy and the purely span-based strategy to the limitations of using span matching, in which relevant values cannot be always found directly in the dialog context.
DS-DST outperforms BERT-DST, which separately encodes dialog context and domain-slot pairs based on BERT,  by $7.81\%$ on MultiWOZ 2.1. The above results shows the effectiveness of our model design based on BERT, where we enforce the strong interactions between dialog context and domain-slot pairs. 

To further investigate the differences and importance of strong interactions, we reproduce the original BERT-DST model described in \cite{rastogi2019towards}. In addition, we compare with ToD-BERT \cite{wu2020tod}, which is a large pre-trained model based on several task-oriented dialogue datasets, and it also separately encodes dialog context and domain-slot pairs. We show the results in Table \ref{table-compare-with-google}.~\footnote{Here we did not show the results when treating all the slots as non-categorical slots, one reason is that the performances of BERT-DST-Span are much worse than BERT-DST.} We observe that our model is consistently much better than BERT-DST and BERT-DST-Picklist. Moreover, our models based on BERT surpass the strong ToD-BERT.
% When having access to the full ontology, Picklist-DST outperforms Bert-DST-Picklist (whole dialog history) by $6.88\%$.
 We conclude that our improvements come from the strong interactions between slots and dialog context. Therefore, it is important to employ strong interactions to multi-domain DST tasks.

\subsection{Per Slot Accuracy}\label{per-slot-section}
Now that we have observed that DS-DST and DS-Picklist perform much better than DS-Span, we focus on where the accuracy improvement comes from.
Table~\ref{table:per-slot-acc} shows the accuracy for each slot type on the MultiWOZ 2.1 test set,
and we can observe significant improvement over the DS-Span baseline for some slots, including \textit{hotel-type}, \textit{attraction-type}, \textit{attraction-name}, \textit{hotel-internet} and \textit{hotel-parking}.
This is because their values usually have different expressions and cannot be extracted from the dialog context, which decreases the performance of the span-based methods.
In contrast, their values can be predicted directly from the candidate-value lists. Compared with other slots, these slots still have space for improvements. 

%We observed a consistent trend on WOZ 2.0, and more deitaled are described in Appendix.

%After applying our dual strategy, the slot accuracies of most picklist-based slots are improved.%, and almost half of them are significantly improved. 
%For example, the slot accuracies of \textit{hotel-type} and \textit{hotel-internet} increase by $6.05\%$ and $4.50\%$, respectively.
%This can be ascribed to the fact that their values can be predicted directly from the candidate-value lists. 
%When the full ontology is accessible, DS-Picklist improves the performance of span-based slots, which results in a higher joint accuracy than DS-DST.

% \begin{table}[H]
% \centering
% \begin{tabular}{llll}
% \hline
%                     & Concatenation & Slot Description & Question \\ \hline
% DialogBERT-Default  &    50.60\%           &     50.22\%             &   48.82\%       \\
% DialogBERT-Picklist &      53.30\%         &       52.74\%           &     52.90\%     \\ \hline
% \end{tabular}

\subsection{Analysis and Discussions}

\paragraph{Error analysis}
To better understand the improvement, we conducted an error analysis and inspected actual examples on the MultiWOZ 2.1 validation set.
Table~\ref{table-error-analysis} shows the top-10 slots, according to the ratio of ground-truth slot values which cannot be found through span matching.
That is, for such examples, DS-Span cannot extract the ground-truth strings, resulting in the low joint accuracy.
Here, we show how well our DS-DST and DS-Picklist can correctly predict the missing values in DS-Span.
As we can see in this table, the two methods dramatically reduce the errors for some slots such as \textit{attraction-type}, \textit{hotel-internet} and \textit{hotel-parking}.
Hence, for these kinds of slots, it is better to treat them as categorical slots.
%In Table~\ref{table-error-analysis}, we show an error analysis for the Top-10 slots of the MultiWOZ 2.1 validation set according to the failure ratios of slot values predictions.
%which cannot be found in the dialog context.
%For these slots, DS-Span cannot predict them correctly through span matching in the context.%as the values cannot be predicted directly through span matching in the context. 
%While for those picklist-based slots such as \textit{hotel-type}, \textit{hotel-parking} and \textit{attraction-type}, DS-DST and DS-Picklist can correctly predict the values from the candidate-value lists and reduce the error dramatically. %to a large extent. 
% For the \textit{time} slots such as \textit{taxi-leave at} and \textit{train-arrive by}, DS-DST has high error rates as they are treated as span-based slots, and only few values (i.e. `none' and `dontcare') can be correctly predicted by the slot-gate classification.
Among the top-10 slots, the \textit{time}-related slots such as \textit{taxi-leave at} and \textit{train-arrive by}, which are span-based slots in DS-DST, DS-Span and DS-DST cannot perform well as there are no span matching in the dialogue context, and only few values (i.e., `\textit{none}' and `\textit{dontcare}') can be correctly predicted by the slot-gate classification. 
When the ontology is accessible, DS-Picklist can further reduce the error rates, since the predicted values can be found in the candidate-values lists.  

On the other hand, we also investigated slots whose ground-truth values can be found through span matching, and we did not observe a significant difference between the three methods.
This means that both the non-categorical and categorical methods perform similarly when target values are explicitly mentioned in the dialogues.
Therefore, when most of the slot values can be found directly in the dialog context, these slots can be treated as either non-categorical slots or categorical slots.

%On the other hand, we also investigate the Top-10 slots based on the ratios of slot values which can be found in the slot context. DS-Span, DS-DST and DS-Picklist have similar performance and retain very low error rates. We conjecture that for the slot values which can be found in the dialog context, our models could achieve similar high performances either they are treated as span-based slots or picklist-based slots. 

As our model relies on the slot-gate classification in Section \ref{slot-gate-classify}, we also investigate the potential influence of this module. We replace this module with an oracle slot-gate classification module, and the joint accuracy is improved from $55.23\%$ to $86.10\%$ on the development set of MultiWOZ 2.1, which indicates that there is a great space to improve the performance with better designs of the slot-gate classification module.

\begin{table*}[]
\centering
%\resizebox{1.0\linewidth}{!}{
\begin{adjustbox}{width=\textwidth}
\begin{tabular}{ll}
\hline
\textbf{User}          & i am looking for an expensive place to stay on the north side of cambridge .                                                                                                                                                                                                                                                                    \\
\textbf{System}        & i am sorry , i haven ' t found any matches , would you like me to look for something else ?                                                                                                                                                                                                                                                     \\
\textbf{User}          & i am looking for a 4 star hotel and i need free internet and parking .                                                                                                                                                                                                                                                                          \\ \hdashline
\textbf{Ground Truths} & \begin{tabular}[c]{@{}l@{}}\textless{}hotel, internet, yes\textgreater{}, \textless{}hotel, stars, 4\textgreater{}, \textless{}hotel, parking, yes\textgreater{}, \textless{}hotel, type, hotel\textgreater{}, \textless{}hotel, area, north\textgreater{}, \\ \textless{}hotel, price range, expensive\textgreater{}\end{tabular}              \\
\textbf{DS-Span}       & \begin{tabular}[c]{@{}l@{}}\textbf{\textless{}hotel, internet, free internet\textgreater{}}, \textless{}hotel, stars, 4\textgreater{}, \textbf{\textless{}hotel, parking, internet\textgreater{}}, \textbf{\textless{}hotel, type, none\textgreater{}}, \textless{}hotel, area, north\textgreater{},\\ \textless{}hotel, price range, expensive\textgreater{}\end{tabular} \\
\textbf{DS-DST}        & \begin{tabular}[c]{@{}l@{}}\textless{}hotel, internet, yes\textgreater{}, \textless{}hotel, stars, 4\textgreater{}, \textless{}hotel, parking, yes\textgreater{}, \textbf{\textless{}hotel, type, none\textgreater{}}, \textless{}hotel, area, north\textgreater{}, \\ \textless{}hotel, price range, expensive\textgreater{}\end{tabular}               \\ \hline
\textbf{User}          & it's so hot today , can you help me find a good pool to visit on the north side of the city ?                                                                                                                                                                                                                                                   \\
\textbf{System}        & i have 2 pools in the north area of town : jesus green outdoor pool and kings hedges learner pool . which do you prefer ?                                                                                                                                                                                                                       \\
\textbf{User}          & kings hedges sounds nice . can i get the address please ?                                                                                                                                                                                                                                                                                       \\ \hdashline
\textbf{Ground Truths} & \textless{}attraction, area, north\textgreater{}, \textless{}attraction, type, swimming pool\textgreater{}, \textless{}attraction, name, kings hedges learner pool\textgreater{}                                                                                                                                                                \\
\textbf{DS-Span}       & \textless{}attraction, area, north\textgreater{}, \textbf{\textless{}attraction, type, pool\textgreater{}}, \textless{}attraction, name, kings hedges learner pool\textgreater{}                                                                                                                                                                         \\
\textbf{DS-DST}        & \textless{}attraction, area, north\textgreater{}, \textless{}attraction, type, swimming pool\textgreater{}, \textless{}attraction, name, kings hedges learner pool\textgreater{}                                                                                                                                                                \\ \hline
\textbf{User}          & do you happen to know of any trains leaving for cambridge this wednesday ?                                                                                                                                                                                                                                                                      \\
\textbf{System}        & yes . there are a total of 202 trains leaving for cambridge on wednesday . where will you be departing from ?                                                                                                                                                                                                                                   \\
\textbf{User}          & i will be leaving from norwich and i need to arrive by 8 : 15 .                                                                                                                                                                                                                                                                                 \\
\textbf{System}        & the tr4203 is leaving from norwich to cambridge at 05 : 16 . would you like to book a ticket ?                                                                                                                                                                                                                                                  \\
\textbf{User}          & sure , can you book that for 2 people and provide my reference number ?                                                                                                                                                                                                                                                                         \\ \hdashline
\textbf{Ground Truths} & \begin{tabular}[c]{@{}l@{}}\textless{}train, arrive by, 08 : 15\textgreater{}, \textless{}train, departure, norwich\textgreater{}, \textless{}train, day, wednesday\textgreater{}, \textless{}train, book people, 2\textgreater{}, \\ \textless{}train, destination, cambridge\textgreater{}\end{tabular}                                       \\
\textbf{DS-Span}       & \begin{tabular}[c]{@{}l@{}}\textbf{\textless{}train, arrive by, 8 : 15\textgreater{}}, \textless{}train, departure, norwich\textgreater{}, \textless{}train, day, wednesday\textgreater{}, \textless{}train, book people, 2\textgreater{}, \\ \textless{}train, destination, cambridge\textgreater{}\end{tabular}                                        \\
\textbf{DS-DST}        & \begin{tabular}[c]{@{}l@{}}\textbf{\textless{}train, arrive by, 8 : 15\textgreater{}}, \textless{}train, departure, norwich\textgreater{}, \textless{}train, day, wednesday\textgreater{}, \textless{}train, book people, 2\textgreater{}, \\ \textless{}train, destination, cambridge\textgreater{}\end{tabular}                                        \\
\textbf{DS-Picklist}   & \begin{tabular}[c]{@{}l@{}}\textless{}train, arrive by, 08 : 15\textgreater{}, \textless{}train, departure, norwich\textgreater{}, \textless{}train, day, wednesday\textgreater{}, \textless{}train, book people, 2\textgreater{}, \\ \textless{}train, destination, cambridge\textgreater{}\end{tabular} \\ \hline
\end{tabular}
\end{adjustbox}
\caption{Predicted dialog states on the MultiWOZ 2.1 validation set, bold face means incorrect prediction. The first two examples show comparisons between DS-Span and DS-DST. The last example shows comparisons between DS-Span, DS-DST and DS-Picklist. } \label{table-case-study}
\end{table*}

\paragraph{Examples}
Table \ref{table-case-study} shows three examples of dialogue turns in the validation set.
In the first example, we can see that DS-Span cannot correctly extract the ground-truth values, because the User does not always explicitly mention `\textit{yes}' or `\textit{no}' when being asked about the internet or parking requests.
In the second example, the User and the System are talking about a swimming pool, but they just say `\textit{pool}' and its meaning can be inferred from the context.
As a result, DS-Span can only extract `\textit{pool}' as a value, which is not sufficient.
In the third example, all the predictions are semantically correct; however, in terms of the string match, only DS-Picklist can correctly predict the value.
The two other methods rely on span extraction.
This is caused by formatting issues; that is, it is not always guaranteed that strings in the context satisfy desired formats, such as time expressions.
Based on our analysis, future work needs to consider more relevant evaluation metrics than the widely-used string matching metric.
For example, in the QA research community, it is investigated how to more robustly evaluate QA models~\citep{qa-eval}.

\paragraph{Open discussions}

Multi-domain dialog state tracking is enjoying popularity in enhancing research on task-oriented dialog systems, to handle tasks across different domains and support a large number of services.
However, it should be noted that there is much room for improvement with the popular MultiWOZ 2.0~\cite{budzianowski2018multiwoz} and MultiWOZ 2.1~\cite{eric2019multiwoz} datasets, due to their annotation errors, ambiguity, and inconsistency.
Moreover, a potential problem is that no standard ways have been established for the evaluation of the MultiWOZ dataset.
Some papers are following the pre-processing ways provided by \citet{wu2019transferable}, while others have their own ways, which may result in unfair comparisons; for example, there are some `\textit{none}' values in the test set, and an evaluation metric without considering them will lead to higher accuracy (up to $3\%$ in our experience).
%Recent work has conducted re-annotation of the validation and test sets of the MultiWOZ dataset~\cite{campagna2020state}, or updating the existing datasets to MultiWOZ 2.2 with higher quality and consistency \cite{zang2020multiwoz}.
Recent work has refined the datasets to form the latest MultiWOZ 2.2 dataset with higher quality and consistency~\cite{zang2020multiwoz}.
We encourage providing more details of the data processing in future work, and more importantly, testing models on the test set with the higher quality.

\section{Conclusion} \label{conclusion}

In this paper, we have proposed a dual strategy model with strong interactions between the dialog context and  domain-slot pairs for the task of multi-domain dialog state tracking.
In particular, we predict the slot value via selecting over a partial ontology for categorical slots or finding values from the dialog context for non-categorical slots. Our models achieve the state of the art results on the MultiWOZ 2.0 and competitive results on the MultiWOZ 2.1. Moreover, we conduct a comprehensive analysis on the dataset to facilitate future research.

% In particular, we first treat domain-slot pairs as span-based slots and picklist-based slots, and then we predict the slot value via selecting over a partial ontology (for picklist-based slots) or finding values from the dialog context (for span-based slots).
% It's worth mentioning that our dual strategy is consistent with real scenarios and flexible to real applications and datasets.
% In addition, DS-DST mitigates the issues existing in previous work and achieves state-of-art results on the MultiWOZ 2.0 dataset and MultiWOZ 2.1 dataset. 

\section*{Acknowledgments}
This work is supported in part by NSF under grants III-1763325, III-1909323, and SaTC-1930941. We thank Salesforce research members for their
insightful discussions, and the anonymous reviewers for their
helpful and valuable comments.
%%%%%%%%%%%%%%%%%%%%%%%%%%%%%%%%%

%\bibliography{reference,acl2020}
\bibliography{reference}
\bibliographystyle{acl_natbib}

\clearpage
\newpage

\appendix 
\section{Appendix}

\subsection{Training Details}
We employ a pre-trained BERT model with the ``bert-base-uncased'' configuration.\footnote{\url{https://github.com/huggingface/transformers/tree/master/examples}}
During the fine-tuning process, we update all the model parameters using the  BertAdam~\cite{devlin2018bert} optimizer. The maximum input sequence length after WordPiece tokenization for BERT is set to $512$.
The constant margin $\lambda$ is set to $0.5$ for the DS-DST and DS-Picklist models.
For the MultiWOZ 2.0 and MultiWOZ 2.1 datasets, the initial learning rate is set to $1e-4$, and the proportion for learning rate warmup is set to $0.1$.
The batch size is set to $16$.
We follow the learning rate decay mechanism as in~\cite{lee2019sumbt}, and we set the maximum number of the training epochs to $5$. 
We check the joint accuracy on the validation set after every 1,000 iterations, and keep the best checkpoint to be used for the final evaluation on the test set.

\subsection{Sample Output}

We present outputs of DS-Span and DS-DST in all the turns for two dialogues (i.e., \textit{MUL0729}, \textit{PMUL2428}) on the validation set of the MultiWOZ 2.1. Table~\ref{example-1} and Table~\ref{example-2} show the predicted dialog states for \textit{MUL0729} and \textit{PMUL2428}, respectively. In Table~\ref{example-1}, \textit{hotel type} and \textit{hotel internet} are predicted incorrectly by DS-Span, where the value \textit{yes} of \textit{hotel internet} has a different description \textit{free wifi} in the dialog context. For this type of values, DS-Span cannot find the spans directly in the dialog context. 
In Table~\ref{example-2}, DS-Span does not correctly predict the state \textit{<taxi, departure, funky fun house>} at the $6_{th}$ turn.

\begin{table*}%[H]
\centering
\resizebox{1.0\textwidth}{!}{%
\begin{tabular}{lll}
\hline
                                  & \textbf{User:}                              & i am looking for a train from ely to cambridge . do you have such trains ?                                                                                                                                                                                                                                                                                                                                                                                                                                                                                                               \\
                                  & \textbf{Ground Truths:}                     & {\color[HTML]{333333} \textless{}train, destination, cambridge\textgreater{}, \textless{}train, departure, ely\textgreater{}}                                                                                                                                                                                                                                                                                                                                                                                                                                                            \\
                                  & \textbf{DS-Span:}            & \textless{}train, destination, cambridge\textgreater{}, \textless{}train, departure, ely\textgreater{}                                                                                                                                                                                                                                                                                                                                                                                                                                                                                   \\
\multirow{-4}{*}{\textbf{Turn 1}} & \textbf{DS-DST:}   & \textless{}train, destination, cambridge\textgreater{}, \textless{}train, departure, ely\textgreater{}                                                                                                                                                                                                                                                                                                                                                                                                                                                                                   \\ \hline
                                  & \textbf{System:}                            & there are 70 trains do you have a date and time you would like to go ?                                                                                                                                                                                                                                                                                                                                                                                                                                                                                                                   \\
                                  & \textbf{User:}                              & i would like to leave on a sunday and arrive by 11:30 .                                                                                                                                                                                                                                                                                                                                                                                                                                                                                                                                  \\
                                  & \textbf{Ground Truths:}                     & \textless{}train, arrive by, 11:30\textgreater{}, \textless{}train, destination, cambridge\textgreater{}, \textless{}train, departure, ely\textgreater{}, \textless{}train, day, sunday\textgreater{}                                                                                                                                                                                                                                                                                                                                                                                    \\
                                  & \textbf{DS-Span:}            & \textless{}train, arrive by, 11:30\textgreater{}, \textless{}train, destination, cambridge\textgreater{}, \textless{}train, departure, ely\textgreater{}, \textless{}train, day, sunday\textgreater{}                                                                                                                                                                                                                                                                                                                                                                                    \\
\multirow{-5}{*}{\textbf{Turn 2}} & \textbf{DS-DST:}   & \textless{}train, arrive by, 11:30\textgreater{}, \textless{}train, destination, cambridge\textgreater{}, \textless{}train, departure, ely\}, \textless{}train, day, sunday\textgreater{}                                                                                                                                                                                                                                                                                                                                                                                                \\ \hline
                                  & \textbf{System:}                            & tr4212 is arriving at 9:52 would that work for you ?                                                                                                                                                                                                                                                                                                                                                                                                                                                                                                                                     \\
                                  & \textbf{User:}                              & what time does that train depart ?                                                                                                                                                                                                                                                                                                                                                                                                                                                                                                                                                       \\
                                  & \textbf{Ground Truths:}                     & \textless{}train, arrive by, 11:30\textgreater{}, \textless{}train, destination, cambridge\textgreater{}, \textless{}train, departure, ely\textgreater{}, \textless{}train, day, sunday\textgreater{}                                                                                                                                                                                                                                                                                                                                                                                    \\
                                  & \textbf{DS-Span:}            & \textless{}train, arrive by, 11:30\textgreater{}, \textless{}train, destination, cambridge\textgreater{}, \textless{}train, departure, ely\textgreater{}, \textless{}train, day, sunday\textgreater{}                                                                                                                                                                                                                                                                                                                                                                                    \\
\multirow{-5}{*}{\textbf{Turn 3}} & \textbf{DS-DST:}   & \textless{}train, arrive by, 11:30\textgreater{}, \textless{}train, destination, cambridge\textgreater{}, \textless{}train, departure, ely\textgreater{}, \textless{}train, day, sunday\textgreater{}                                                                                                                                                                                                                                                                                                                                                                                    \\ \hline
                                  & \textbf{System:}                            & the train leaves at 9:35 . shall i book it for you ?                                                                                                                                                                                                                                                                                                                                                                                                                                                                                                                                     \\
                                  & \textbf{User:}                              & no , thank you . i would like to see if you could find a place for me to stay that would have a 4 star rating .                                                                                                                                                                                                                                                                                                                                                                                                                                                                          \\
                                  & \textbf{Ground Truths:}                     & \textless{}train, day, sunday\textgreater{}, \textless{}hotel, stars, 4\textgreater{}, \textless{}train, destination, cambridge\textgreater{}, \textless{}train, arrive by, 11:30\textgreater{}, \textless{}train, departure, ely\textgreater{}                                                                                                                                                                                                                                                                                                                                          \\
                                  & \textbf{DS-Span:}            & \textless{}train, day, sunday\textgreater{}, \textless{}hotel, stars, 4\textgreater{}, \textless{}train, destination, cambridge\textgreater{}, \textless{}train, arrive by, 11:30\textgreater{}, \textless{}train, departure, ely\textgreater{}                                                                                                                                                                                                                                                                                                                                          \\
\multirow{-5}{*}{\textbf{Turn 4}} & \textbf{DS-DST:}   & \textless{}train, day, sunday\textgreater{}, \textless{}hotel, stars, 4\textgreater{}, \textless{}train, destination, cambridge\textgreater{}, \textless{}train, arrive by, 11:30\textgreater{}, \textless{}train, departure, ely\textgreater{}                                                                                                                                                                                                                                                                                                                                          \\ \hline
                                  & \textbf{System:}                            & of course , there are many 4 star hotel -s to choose from . would you like me to book 1 for you ?                                                                                                                                                                                                                                                                                                                                                                                                                                                                                        \\
                                  & \textbf{User:}                              & not yet , thanks . are any of them guesthouses ?                                                                                                                                                                                                                                                                                                                                                                                                                                                                                                                                         \\
                                  & \textbf{Ground Truths:}                     & \textless{}train, day, sunday\textgreater{}, \textless{}hotel, stars, 4\textgreater{}, \textless{}train, destination, cambridge\textgreater{}, \textless{}train, arrive by, 11:30\textgreater{}, \textless{}train, departure, ely\textgreater{}, \textless{}hotel, type, guest house\textgreater{}                                                                                                                                                                                                                                                                                       \\
                                  & \textbf{DS-Span:}            & \textless{}train, day, sunday\textgreater{}, \textless{}hotel, stars, 4\textgreater{}, \textless{}train, destination, cambridge\textgreater{}, \textless{}train, arrive by, 11:30\textgreater{}, \textless{}train, departure, ely\textgreater{}, {\ul \textbf{\textless{}hotel, type, hotel\textgreater{}}}                                                                                                                                                                                                                                                                                                 \\
\multirow{-5}{*}{\textbf{Turn 5}} & \textbf{DS-DST:}   & \textless{}train, day, sunday\textgreater{}, \textless{}hotel, stars, 4\textgreater{}, \textless{}train, destination, cambridge\textgreater{}, \textless{}train, arrive by, 11:30\textgreater{}, \textless{}train, departure, ely\textgreater{}, \textless{}hotel, type, guest house\textgreater{}                                                                                                                                                                                                                                                                                       \\ \hline
                                  & \textbf{System:}                            & there are 18 guesthouses to choose from , do you have a preference to the area you would like to stay ?                                                                                                                                                                                                                                                                                                                                                                                                                                                                                  \\
                                  & \textbf{User:}                              & i need a 4 star , and in the east with free wifi for 4 people , 5 nights . i'll need a reference number .                                                                                                                                                                                                                                                                                                                                                                                                                                                                                \\
                                  & \textbf{Ground Truths:}                     & \begin{tabular}[c]{@{}l@{}}\textless{}train, day, sunday\textgreater{}, \textless{}hotel, book stay, 5\textgreater{}, \textless{}hotel, book people, 4\textgreater{}, \textless{}hotel, stars, 4\textgreater{}, \textless{}train, destination, cambridge\textgreater{}, \textless{}hotel, internet, yes\textgreater{}, \\ \textless{}train, arrive by, 11:30\textgreater{}, \textless{}train, departure, ely\textgreater{}, \textless{}hotel, area, east\textgreater{}, \textless{}hotel, type, guest house\textgreater{}\end{tabular}                                                   \\
                                  & \textbf{DS-Span:}            & \begin{tabular}[c]{@{}l@{}}\textless{}train, day, sunday\textgreater{}, \textless{}hotel, book stay, 5\textgreater{}, \textless{}hotel, book people, 4\textgreater{}, \textless{}hotel, stars, 4\textgreater{}, \textless{}train, destination, cambridge\textgreater{}, {\ul \textbf{\textless{}hotel, internet, no\textgreater{}}}, \\ \textless{}train, arrive by, 11:30\textgreater{}, \textless{}train, departure, ely\textgreater{}, \textless{}hotel, area, east\textgreater{}, {\ul \textbf{\textless{}hotel, type, hotel\textgreater{}}} \end{tabular}                                                               \\
\multirow{-5}{*}{\textbf{Turn 6}} & \textbf{DS-DST:}   & \begin{tabular}[c]{@{}l@{}}\textless{}train, day, sunday\textgreater{}, \textless{}hotel, book stay, 5\textgreater{}, \textless{}hotel, book people, 4\textgreater{}, \textless{}hotel, stars, 4\textgreater{}, \textless{}train, destination, cambridge\textgreater{}, \textless{}hotel, internet, yes\textgreater{}, \\ \textless{}train, arrive by, 11:30\textgreater{}, \textless{}train, departure, ely\textgreater{}, \textless{}hotel, area, east\textgreater{}, \textless{}hotel, type, guest house\textgreater{}\end{tabular}                                                   \\ \hline
                                  & \textbf{System:}                            & do you want that guesthouse reservation to begin on sunday ?                                                                                                                                                                                                                                                                                                                                                                                                                                                                                                                             \\
                                  & \textbf{User:}                              & yes . i need 5 nights starting on sunday .                                                                                                                                                                                                                                                                                                                                                                                                                                                                                                                                               \\
                                  & \textbf{Ground Truths:}                     & \begin{tabular}[c]{@{}l@{}}\textless{}train, day, sunday\textgreater{}, \textless{}hotel, book stay, 5\textgreater{}, \textless{}hotel, book people, 4\textgreater{}, \textless{}hotel, stars, 4\textgreater{}, \textless{}train, destination, cambridge\textgreater{}, \textless{}hotel, internet, yes\textgreater{},\\  \textless{}hotel, type, guest house\textgreater{}, \textless{}train, arrive by, 11:30\textgreater{}, \textless{}train, departure, ely\textgreater{}, \textless{}hotel, area, east\textgreater{}, \textless{}hotel, book day, sunday\textgreater{}\end{tabular} \\
                                  & \textbf{DS-Span:}            & \begin{tabular}[c]{@{}l@{}}\textless{}train, day, sunday\textgreater{}, \textless{}hotel, book stay, 5\textgreater{}, \textless{}hotel, book people, 4\textgreater{}, \textless{}hotel, stars, 4\textgreater{}, \textless{}train, destination, cambridge\textgreater{}, \textless{}hotel, internet, yes\textgreater{}, \\ {\ul \textbf{\textless{}hotel, type, hotel\textgreater{}}}, \textless{}train, arrive by, 11:30\textgreater{}, \textless{}train, departure, ely\textgreater{}, \textless{}hotel area, east\textgreater{}, \textless{}hotel, book day, sunday\textgreater{}\end{tabular}             \\
\multirow{-5}{*}{\textbf{Turn 7}} & \textbf{DS-DST:}   & \begin{tabular}[c]{@{}l@{}}\textless{}train, day, sunday\textgreater{}, \textless{}hotel, book stay, 5\textgreater{}, \textless{}hotel, book people, 4\textgreater{}, \textless{}hotel, stars, 4\textgreater{}, \textless{}train, destination, cambridge\textgreater{}, \textless{}hotel, internet, yes\textgreater{}, \\ \textless{}hotel, type, guest house\textgreater{}, \textless{}train, arrive by, 11:30\textgreater{}, \textless{}train, departure, ely\textgreater{}, \textless{}hotel, area, east\textgreater{}, \textless{}hotel, book day, sunday\textgreater{}\end{tabular}        \\ \hline
                                  & \textbf{System:}                            & i have confirmed your reservation at allenbell starting on sunday for 5 nights . your reference number is sltivabu . can i help with anything else ?                                                                                                                                                                                                                                                                                                                                                                                                                                     \\
                                  & \textbf{User:}                              & that is all i need . thank you so much for all your help .                                                                                                                                                                                                                                                                                                                                                                                                                                                                                                                               \\
                                  & \textbf{Ground Truths:}                     & \begin{tabular}[c]{@{}l@{}}\textless{}train, day, sunday\textgreater{}, \textless{}hotel, book stay, 5\textgreater{}, \textless{}hotel, book people, 4\textgreater{}, \textless{}hotel, stars, 4\textgreater{}, \textless{}train, destination, cambridge\textgreater{}, \textless{}hotel, internet, yes\textgreater{}, \\ \textless{}hotel, type, guest house\textgreater{}, \textless{}train, arrive by, 11:30\textgreater{}, \textless{}train, departure, ely\textgreater{}, \textless{}hotel, area, east\textgreater{}, \textless{}hotel, book day, sunday\textgreater{}\end{tabular} \\
                                  & \textbf{DS-Span:} & \begin{tabular}[c]{@{}l@{}}\textless{}train, day, sunday\textgreater{}, \textless{}hotel, book stay, 5\textgreater{}, \textless{}hotel, book people, 4\textgreater{}, \textless{}hotel, stars, 4\textgreater{}, \textless{}train, destination, cambridge\textgreater{}, \textless{}hotel, internet, yes\textgreater{}, \\ {\ul \textbf{\textless{}hotel, type, hotel\textgreater{}}}, \textless{}train, arrive by, 11:30\textgreater{}, \textless{}train, departure, ely\textgreater{}, \textless{}hotel, area, east\textgreater{}, \textless{}hotel, book day, sunday\textgreater{}\end{tabular}           \\
\multirow{-5}{*}{\textbf{Turn 8}} & \textbf{DS-DST:}   & \begin{tabular}[c]{@{}l@{}}\textless{}train, day, sunday\textgreater{}, \textless{}hotel, book stay, 5\textgreater{}, \textless{}hotel, book people, 4\textgreater{}, \textless{}hotel, stars, 4\textgreater{}, \textless{}train, destination, cambridge\textgreater{}, \textless{}hotel, internet, yes\textgreater{}, \\ \textless{}hotel, type, guest house\textgreater{}, \textless{}train, arrive by, 11:30\textgreater{}, \textless{}train, departure, ely\textgreater{}, \textless{}hotel, area, east\textgreater{}, \textless{}hotel, book day, sunday\textgreater{}\end{tabular} \\ \hline
\end{tabular}}\caption{Predicted dialog states of DS-Span and DS-DST for domains (i.e., \textit{train}, \textit{hotel}) in dialogue \textit{MUL0729} from the MultiWOZ 2.1 dataset. }\label{example-1}
\end{table*}

%%%%%%%%%%%%%%%%%%%%%%%%%

% \usepackage{multirow}
% \usepackage{multirow}

\begin{table*}%[H]
\centering
% \resizebox{0.93\linewidth}{!}{%
\resizebox{1.0\textwidth}{!}{%
\begin{tabular}{lll}
\hline
                                  & \textbf{User:}                            & i am planning a trip to go to a particular restaurant , can you assist ?                                                                                                                                                                                                                                                                                                                                                                                                                                                                      \\
                                  & \textbf{Ground Truths:}                   & {\color[HTML]{333333} }                                                                                                                                                                                                                                                                                                                                                                                                                                                                                                                       \\
                                  & \textbf{DS-Span:}          &                                                                                                                                                                                                                                                                                                                                                                                                                                                                                                                                               \\
\multirow{-4}{*}{\textbf{Turn 1}} & \textbf{DS-DST:} &                                                                                                                                                                                                                                                                                                                                                                                                                                                                                                                                               \\ \hline
                                  & \textbf{System:}                          & sure , what is the name of this particular restaurant ?                                                                                                                                                                                                                                                                                                                                                                                                                                                                                       \\
                                  & \textbf{User:}                            & it s called nandos and i would like to book it for monday at 15:00 . there will be 6 people .                                                                                                                                                                                                                                                                                                                                                                                                                                                 \\
                                  & \textbf{Ground Truths:}                   & \textless{}restaurant, book day, monday\textgreater{}, \textless{}restaurant, name, nandos\textgreater{}, \textless{}restaurant, book time, 15:00\textgreater{}, \textless{}restaurant, book people, 6\textgreater{}                                                                                                                                                                                                                                                                                                                          \\
                                  & \textbf{DS-Span:}          & \textless{}restaurant, book day, monday\textgreater{}, \textless{}restaurant, name, nandos\textgreater{}, \textless{}restaurant, book time, 15:00\textgreater{}, \textless{}restaurant, book people, 6\textgreater{}                                                                                                                                                                                                                                                                                                                          \\
\multirow{-5}{*}{\textbf{Turn 2}} & \textbf{DS-DST:} & \textless{}restaurant, book day, monday\textgreater{}, \textless{}restaurant, name, nandos\textgreater{}, \textless{}restaurant, book time, 15:00\textgreater{}, \textless{}restaurant, book people, 6\textgreater{}                                                                                                                                                                                                                                                                                                                          \\ \hline
                                  & \textbf{System:}                          & no problem ! i have your table reserved for 15:00 on monday . they will hold your table for 15 minutes , your reference number is hvb51vam .                                                                                                                                                                                                                                                                                                                                                                                                  \\
                                  & \textbf{User:}                            & thank you . am also looking for place -s to go in town . the attraction should be in the east                                                                                                                                                                                                                                                                                                                                                                                                                                                 \\
                                  & \textbf{Ground Truths:}                   & \textless{}restaurant, book day, monday\textgreater{}, \textless{}restaurant, name, nandos\textgreater{},
                                  \textless{}restaurant, book people, 6\textgreater{},
                                  \textless{}attraction, area, east\textgreater{},  \textless{}restaurant, book time, 15:00\textgreater{}                                                                                                                                                                                                                                                                         \\
                                  & \textbf{DS-Span:}          & \textless{}restaurant, book day, monday\textgreater{}, \textless{}restaurant, name, nandos\textgreater{}, \textless{}restaurant, book people, 6\textgreater{}, \textless{}attraction, area, east\textgreater{}, \textless{}restaurant, book time, 15:00\textgreater{}                                                                                                                                                                                                                                                                         \\
\multirow{-5}{*}{\textbf{Turn 3}} & \textbf{DS-DST:} & \textless{}restaurant, book day, monday\textgreater{}, \textless{}restaurant, name, nandos\textgreater{}, \textless{}restaurant, book people, 6\textgreater{}, \textless{}attraction, area, east\textgreater{}, \textless{}restaurant, book time, 15:00\textgreater{}                                                                                                                                                                                                                                                                         \\ \hline
                                  & \textbf{System:}                          & what are you looking to do ? my system has located 10 place -s .                                                                                                                                                                                                                                                                                                                                                                                                                                                                              \\
                                  & \textbf{User:}                            & could you recommend something ?                                                                                                                                                                                                                                                                                                                                                                                                                                                                                                               \\
                                  & \textbf{Ground Truths:}                   & \textless{}restaurant, book day, monday\textgreater{}, \textless{}restaurant, name, nandos\textgreater{},  \textless{}restaurant, book people, 6\textgreater{},
                                  \textless{}attraction, area, east\textgreater{},
                                  \textless{}restaurant, book time, 15:00\textgreater{}                                                                                                                                                                                                                                                                         \\
                                  & \textbf{DS-Span:}          & \textless{}restaurant, book day, monday\textgreater{}, \textless{}restaurant, name, nandos\textgreater{},  \textless{}restaurant, book people, 6\textgreater{},
                                  \textless{}attraction, area, east\textgreater{},
                                  \textless{}restaurant, book time, 15:00\textgreater{}                                                                                                                                                                                                                                                                         \\
\multirow{-5}{*}{\textbf{Turn 4}} & \textbf{DS-DST:} & \textless{}restaurant, book day, monday\textgreater{}, \textless{}restaurant, name, nandos\textgreater{},
\textless{}restaurant, book people, 6\textgreater{},
\textless{}attraction, area, east\textgreater{},  \textless{}restaurant, book time, 15:00\textgreater{}                                                                                                                                                                                                                                                                         \\ \hline
                                  & \textbf{System:}                          & yes , the funky fun house is a great place .                                                                                                                                                                                                                                                                                                                                                                                                                                                                                                  \\
                                  & \textbf{User:}                            & great ! can you tell me what the entrance fee is please ?                                                                                                                                                                                                                                                                                                                                                                                                                                                                 \\
                                  & \textbf{Ground Truths:}  
                                  
                                  & 
                                  \begin{tabular}[c]{@{}l@{}} \textless{}restaurant, book day, monday\textgreater{}, \textless{}restaurant, name, nandos\textgreater{}, \textless{}restaurant, book people, 6\textgreater{}, \textless{}attraction, area, east\textgreater{},  \textless{}restaurant, book time, 15:00\textgreater{}, \\ \textless{}attraction, name, funky fun house\textgreater{} \end{tabular}           
                                  
                                                                                                                                                                                                                                             \\
                                  & \textbf{DS-Span:}          & \begin{tabular}[c]{@{}l@{}} \textless{}restaurant, book day, monday\textgreater{}, \textless{}restaurant, name, nandos\textgreater{}, \textless{}restaurant, book people, 6\textgreater{}, \textless{}attraction, area, east\textgreater{}, \textless{}restaurant, book time, 15:00\textgreater{}, \\ \textless{}attraction, name, funky fun house\textgreater{} \end{tabular}                                                                                                                                                                                                                     \\
\multirow{-5}{*}{\textbf{Turn 5}} & \textbf{DS-DST:} & \begin{tabular}[c]{@{}l@{}}\textless{}restaurant, book day, monday\textgreater{}, \textless{}restaurant, name, nandos\textgreater{}, \textless{}restaurant, book people, 6\textgreater{}, \textless{}attraction, area, east\textgreater{},  \textless{}restaurant, book time, 15:00\textgreater{} \\  \textless{}attraction, name, funky fun house\textgreater{}\end{tabular}                                                                                                                                                                                                            \\ \hline
                                  & \textbf{System:}                          & no , i am sorry . you will have to call them for the entrance fee . the phone number is 01223304705 .                                                                                                                                                                                                                                                                                                                                                                                                                                         \\
                                  & \textbf{User:}                            & thank you . i am also looking to book a taxi to travel between the 2 . i need it to arrive to the restaurant by the reservation time .                                                                                                                                                                                                                                                                                                                                                                                                        \\
                                  & \textbf{Ground Truths:}                   & \begin{tabular}[c]{@{}l@{}}\textless{}restaurant, book day, monday\textgreater{}, \textless{}restaurant, name, nandos\textgreater{},   \textless{}restaurant, book people, 6\textgreater{},  \textless{}attraction, area, east\textgreater{}, \textless{}restaurant, book time, 15:00\textgreater{}, \\ \textless{}taxi, arrive by, 15:00\textgreater{}, 
                                  \textless{}attraction, name, funky fun house\textgreater{}, 
                                  \textless{}taxi, destination, nandos\textgreater{}, 
                                  \textless{}taxi, departure, funky fun house\textgreater{}\end{tabular} \\
                                  & \textbf{DS-Span:}          & \begin{tabular}[c]{@{}l@{}}\textless{}restaurant, book day, monday\textgreater{}, \textless{}restaurant, name, nandos\textgreater{},   \textless{}restaurant, book people, 6\textgreater{},  \textless{}attraction, area, east\textgreater{}, \textless{}restaurant, book time, 15:00\textgreater{}, \\ \textless{}taxi, arrive by, 15:00\textgreater{}, 
                                  \textless{}attraction, name, funky fun house\textgreater{}, 
                                  \textless{}taxi, destination, nandos\textgreater{},
                                  {\ul \textbf{ \textless{}taxi, departure, none\textgreater{}}}\end{tabular}                                                             \\
\multirow{-5}{*}{\textbf{Turn 6}} & \textbf{DS-DST:} & \begin{tabular}[c]{@{}l@{}}\textless{}restaurant, book day, monday\textgreater{}, \textless{}restaurant, name, nandos\textgreater{}, \textless{}restaurant, book people, 6\textgreater{},  \textless{}attraction, area, east\textgreater{},  \textless{}restaurant, book time, 15:00\textgreater{}, \\ \textless{}taxi, arrive by, 15:00\textgreater{}, \textless{}attraction, name, funky fun house\textgreater{} , \textless{}taxi, destination, nandos\textgreater{},  \textless{}taxi, departure, funky fun house\textgreater{}\end{tabular}  \\ \hline
                                  & \textbf{System:}                          & your taxi is booked and will be a white audi . the contact number is 07057575130 . how else may i help you ?                                                                                                                                                                                                                                                                                                                                                                                                                                  \\
                                  & \textbf{User:}                            & that s all . thank you for your help !                                                                                                                                                                                                                                                                                                                                                                                                                                                                                                        \\
                                  & \textbf{Ground Truths:}                   & \begin{tabular}[c]{@{}l@{}}\textless{}restaurant, book day, monday\textgreater{}, \textless{}restaurant, name, nandos\textgreater{}, \textless{}restaurant, book people, 6\textgreater{},  \textless{}attraction, area, east\textgreater{},  \textless{}restaurant, book time, 15:00\textgreater{}, \\ \textless{}taxi, arrive by, 15:00\textgreater{}, \textless{}attraction, name, funky fun house\textgreater{} , \textless{}taxi, destination, nandos\textgreater{},   \textless{}taxi, departure, funky fun house\textgreater{}\end{tabular}  \\
                                  & \textbf{DS-Span:}          & \begin{tabular}[c]{@{}l@{}}\textless{}restaurant, book day, monday\textgreater{}, \textless{}restaurant, name, nandos\textgreater{}, \textless{}restaurant, book people, 6\textgreater{},  \textless{}attraction, area, east\textgreater{},  \textless{}restaurant, book time, 15:00\textgreater{}, \\ \textless{}taxi, arrive by, 15:00\textgreater{}, \textless{}attraction, name, funky fun house\textgreater{} , \textless{}taxi, destination, nandos\textgreater{},  \textless{}taxi, departure, funky fun house\textgreater{}\end{tabular}                                                                                                                                                                                                                       \\
\multirow{-5}{*}{\textbf{Turn 7}} & \textbf{DS-DST:} & \begin{tabular}[c]{@{}l@{}}\textless{}restaurant, book day, monday\textgreater{}, \textless{}restaurant, name, nandos\textgreater{}, \textless{}restaurant, book people, 6\textgreater{},  \textless{}attraction, area, east\textgreater{},  \textless{}restaurant, book time, 15:00\textgreater{}, \\ \textless{}taxi, arrive by, 15:00\textgreater{}, \textless{}attraction, name, funky fun house\textgreater{} , \textless{}taxi, destination, nandos\textgreater{},  \textless{}taxi, departure, funky fun house\textgreater{}\end{tabular}                                                                                                                                                                                                                                                       \\ \hline
\end{tabular}}
\caption{Predicted dialog states of DS-Span and DS-DST for domains (i.e., \textit{taxi}, \textit{attraction}, \textit{restaurant}) in dialogue \textit{PMUL2428} from the MultiWOZ 2.1 dataset. }\label{example-2}
\end{table*}

\end{document}